\pdfoutput=1
\documentclass[11pt]{article}

\usepackage[final]{acl}

\usepackage{times}
\usepackage{latexsym}

\usepackage[T1]{fontenc}

\usepackage[utf8]{inputenc}
\usepackage{pdflscape}

\usepackage{microtype}

\usepackage{inconsolata}

\usepackage{graphicx}
\usepackage{amsmath}
\usepackage{algorithm}
\usepackage{algpseudocode}

\usepackage{times}
\usepackage{booktabs}
\usepackage{multirow}
\usepackage{soul}
\usepackage{pgfplots}
\usepackage{hyperref}
\usepackage{url}
\usepackage[most]{tcolorbox}
\usepackage{multicol}

\usepackage{physics}
\usepackage{verbatim}
\usepackage{amssymb}
\usepackage{pifont}
\usepackage{ulem}
\usepackage{mathrsfs} 
\usepackage{fancyvrb}
\pgfplotsset{compat=1.18} 
\useunder{\uline}{\ul}{}
\usepackage{colortbl}
\PassOptionsToPackage{prologue,dvipsnames}{xcolor}

\definecolor{codegreen}{rgb}{0,0.6,0}
\definecolor{codegray}{rgb}{0.5,0.5,0.5}
\definecolor{codepurple}{rgb}{0.58,0,0.82}
\definecolor{backcolour}{rgb}{0.95,0.95,0.92}
\usepackage{listings}

\lstdefinestyle{mystyle}{
  commentstyle=\color{codegreen},
  keywordstyle=\color{magenta},
  numberstyle=\tiny\color{codegray},
  stringstyle=\color{codepurple},
  basicstyle=\ttfamily\footnotesize,
  breakatwhitespace=false,         
  breaklines=true,                 
  captionpos=b,                    
  keepspaces=false,                                 
  showspaces=false,                
  showstringspaces=false,
  showtabs=false,                  
  tabsize=2
}

\title{Control Large Language Models via Divide and Conquer}

\author{
Bingxuan Li$^{\dagger}$ \ \ \ \ Yiwei Wang$^{\mathsection\dagger}$ \ \ \ \ Tao Meng$^{\dagger}$  \ \ \ \
Kai-Wei Chang$^{\dagger}$\ \ \ \ Nanyun Peng$^{\dagger}$ \\ 
$^\dagger$University of California, Los Angeles \quad $^\mathsection$ University of California, Merced \\
\texttt{bingxuan@ucla.edu}
}

\begin{document}
\maketitle
    
\begin{abstract}

This paper investigates controllable generation for large language models (LLMs) with prompt-based control, focusing on Lexically Constrained Generation (LCG).
We systematically evaluate the performance of LLMs on satisfying lexical constraints with prompt-based control, as well as their efficacy in downstream applications. We conclude that LLMs face significant challenges in consistently satisfying lexical constraints with prompt-based control. We identified three key limitations of LLMs for LCG, including (1) position bias, where LLMs tend to satisfy constraints that appear in specific positions within the input; (2) low responsiveness to decoding parameters, which render minimal impact on control of LLMs; and (3) struggle with handling the inherent complexity of certain constraints (e.g., compound words). 
To address these issues, we introduce a Divide and Conquer Generation strategy, effective for both white-box and black-box LLMs, to enhance LLMs performance in LCG tasks, which demonstrates over 90\% improvement on success rate in the most challenging LCG task. Our analysis provides valuable insights into the performance of LLMs in LCG with prompt-based control, and our proposed strategy offers a pathway to more sophisticated and customized text generation applications.

\end{abstract}

\section{Introduction}

\begin{figure}[ht]
    \centering
    \includegraphics[width=\linewidth]{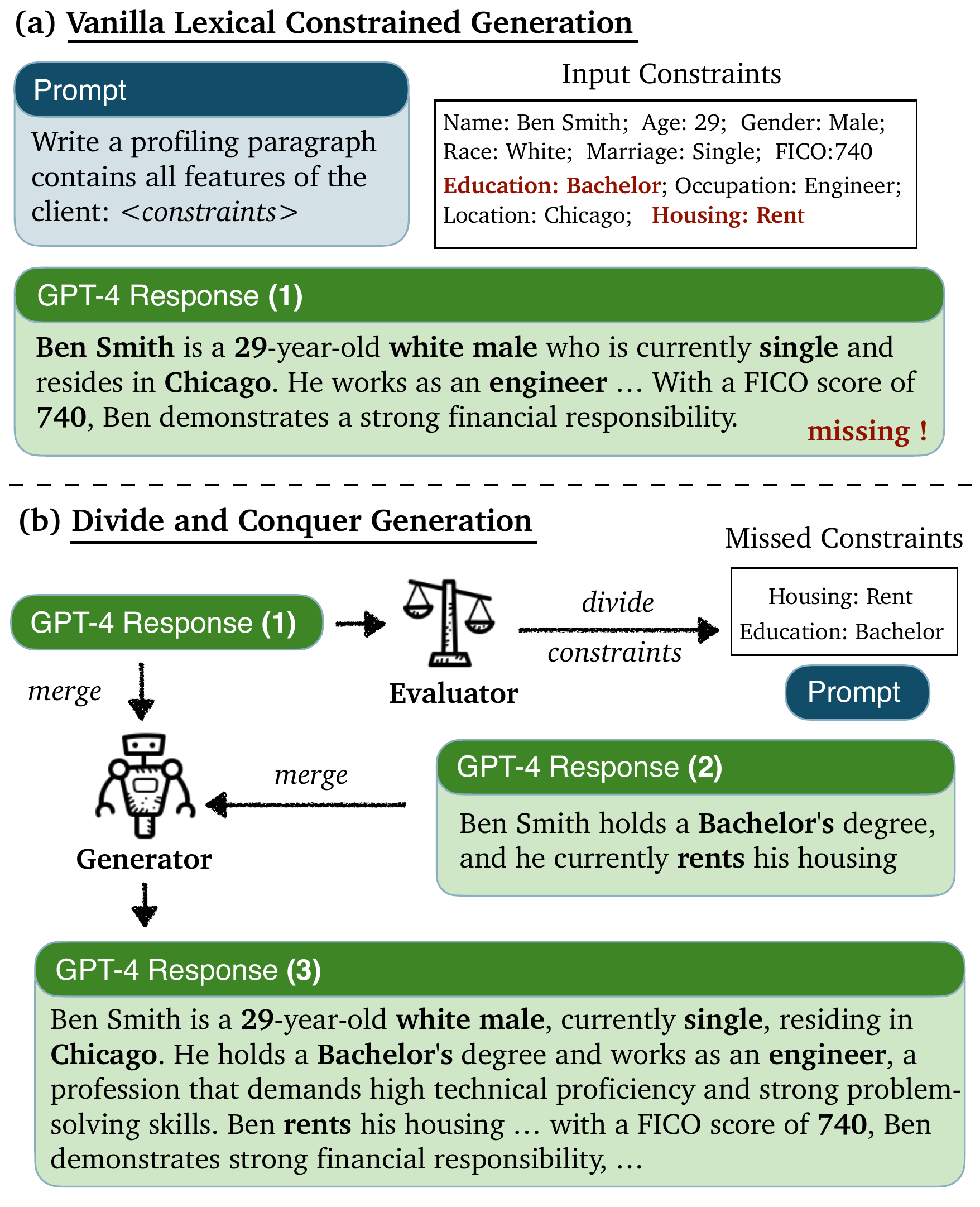}
    \caption{Sub-figure (a) illustrates that modern LLMs struggle to consistently meet complex lexical constraints in real-world scenarios. As shown in sub-figure (b), the Divide and Conquer Generation strategy divides the constraints into two parts (satisfied and missed), then generates a response with the missed constraints - i.e. response (2) - and merges them with the satisfied ones, enhancing the LLMs' ability to meet all specified constraints.}
    \label{fig:teaser}
\end{figure}

Lexically Constrained Generation (LCG) is a crucial task of text generation \cite{zhang2023survey}. By enforcing the inclusion of pre-specified words in the output, LCG facilitates the generation of more faithful and relevant texts. It is helpful for various real-world applications, such as story generation \cite{yao2019plan,goldfarb2020content}, poetry composition~\cite{tian2022sonnet} table-to-text generation \cite{chen2023towards}, and recipe generation \cite{h2020recipegpt}. 

To generate text that adheres to lexical constraint effectively, previous works either design constrained decoding strategies, develop specialized model structures, or present refined mechanisms \cite{sha-2020-gradient, lu2021neurologic, qian2022controllable, meng2022controllable,zhang2023gelato,lu2024nado2,zhang2024adaptable}. However, these approaches often come with significant drawbacks, such as high inference times, complex implementations, requirement of access to output distributions, and unstable text quality. 

Recent advancements in pretrained large language models (LLMs) have showcased their robust few-shot capabilities \cite{brown2020language,ouyang2022training,achiam2023gpt}. Instruction tuning has further enhanced LLMs' ability to generate text that meets controllable constraints as desired by humans \cite{zhang2023instruction}. These developments make prompt-based control an increasingly efficient and practical method for tackling LCG tasks \cite{yang2022tailor}. Notably, prompt-based control has shown markedly strength and robustness compared to earlier methods for LCG \cite{sun-etal-2023-evaluating, ashok2024controllable}, which motivate us to ask: \textit{With prompt-based control, can LLMs consistently satisfy lexical constraints?}

Many recent works investigating prompt-based control of LLMs \cite{sun-etal-2023-evaluating, zhang2023survey, ashok2024controllable} concluded that LLMs have shown effectiveness in satisfying lexical constraints. However, their experiments have typically involved relatively simple tasks with a narrow scope. This leaves a significant gap in detailed understanding of their proficiency, limitations when it comes to satisfying lexical constraints, and effectiveness in real-world applications.

To address this gap, we present a systematic analysis of LLMs' performance in generating text under lexical constraints. We also evaluate their utility in downstream applications where adhering to specific keywords constraints is crucial. Through extensive experiments, we conclude that \textbf{LLMs struggle to adapt to increasingly complex lexical constraints with prompt-based control}. We confirm an intuitive complexity bottleneck: As the number of constraints increases, LLMs' performance \textit{decreases dramatically}. We make several complementary observations: 
\begin{enumerate}
\item Position Bias: The position of each constraint within the prompt can substantially influence the model’s output.
\item Low responsiveness to control decoding parameters: Decoding parameters are not highly sensitive for LLMs in LCG task, especially for temperature and top-k.
\item Inherent Complexity of compound words as constraints: LLMs tend to break down compound words into subwords, which can lead to misinterpretations or alteration of the intended meaning of the output significantly.
\end{enumerate}

To tackle the observed challenges, we introduce an effective strategy - Divide and Conquer Generation - to enhance the ability of models to meet lexical constraints and significantly improve performance. This also helps LLMs achieve more satisfying results in downstream applications. Notably, the Divide and Conquer Generation strategy enables LLaMA-7b to improve the success rate by 93\% in the most challenging LCG task, about 40\% higher than the baseline. Our method is well-suited for both white-box and black-box models, making it an invaluable tool for a broad scope of applications across diverse modeling environments.

Overall, our research conducts an in-depth analysis of LLMs in satisfying lexical constraints, identifies the current challenges faced by LLMs in satisfying lexical constraints, and provides a viable solution to these challenges, paving the way for more sophisticated downstream applications.

\section{Lexically Constrained Generation}

\subsection{Task Setup}
Following previous works \cite{lin2019commongen, zhou2023controlled}, we refer to constraints that require the generated text to include certain keywords in the output as lexical constraints. We consider an input prompt composed of a series of tokens, containing a set of constraints \(X = [x_1, \ldots, x_m]\), where \(x_i\) represent a keyword that must be included. The target output is a coherent sentence \(Y = [y_1, \ldots, y_N]\), with each \(y_i\) is a token. The task is to map the constraint set \(X\) into an appropriate sentence \(Y\) that both adheres to the prompt's requirements (e.g. generate a recipe) and satisfied the defined constraints(e.g. generate sentence that contain all given keywords) .

\paragraph{Evaluation Metrics} We introduce two evaluation metrics in this study:
\begin{enumerate}
    \item \textbf{Instance Success Rate} \((R_{\text{instance}})\): This metric evaluates whether each generated instance satisfies all specified constraints. It is defined as:
    \[
    R_{\text{instance}}(X, Y) = \begin{cases} 
    1 & \text{if } X \subseteq Y,\\ 
    0 & \text{otherwise.} 
    \end{cases}
    \]

    \item \textbf{Keyword Coverage Rate} \((S_{\text{keyword}})\): This metric measures the proportion of input constraints included in the generated texts. It is calculated as:
    \[
    R_{\text{keyword}} = \frac{\text{Number of Satisfied constraints}}{\text{Total number of constraints }}
    \]
\end{enumerate}

\paragraph{Evaluate with LLMs} We have conducted tests using various language models, including LLaMA2-7b-chat, LLaMA2-13b-chat \cite{touvron2023llama}, LLaMA3-8b-chat \cite{metaMetaLlama}, GPT-3.5\cite{achiam2023gpt}, and GPT-4 \cite{achiam2023gpt}. In these experiments, we tasked the models with generating outputs based on specific constraints. Unless stated otherwise, all experiments in this section utilized a greedy decoding strategy for generating responses. Prompt used in evaluation is attached to Appendix \ref{sec:prompts}.

\subsection{Simple Constraints}

\begin{figure}[t]
\centering
\includegraphics[width=\columnwidth]{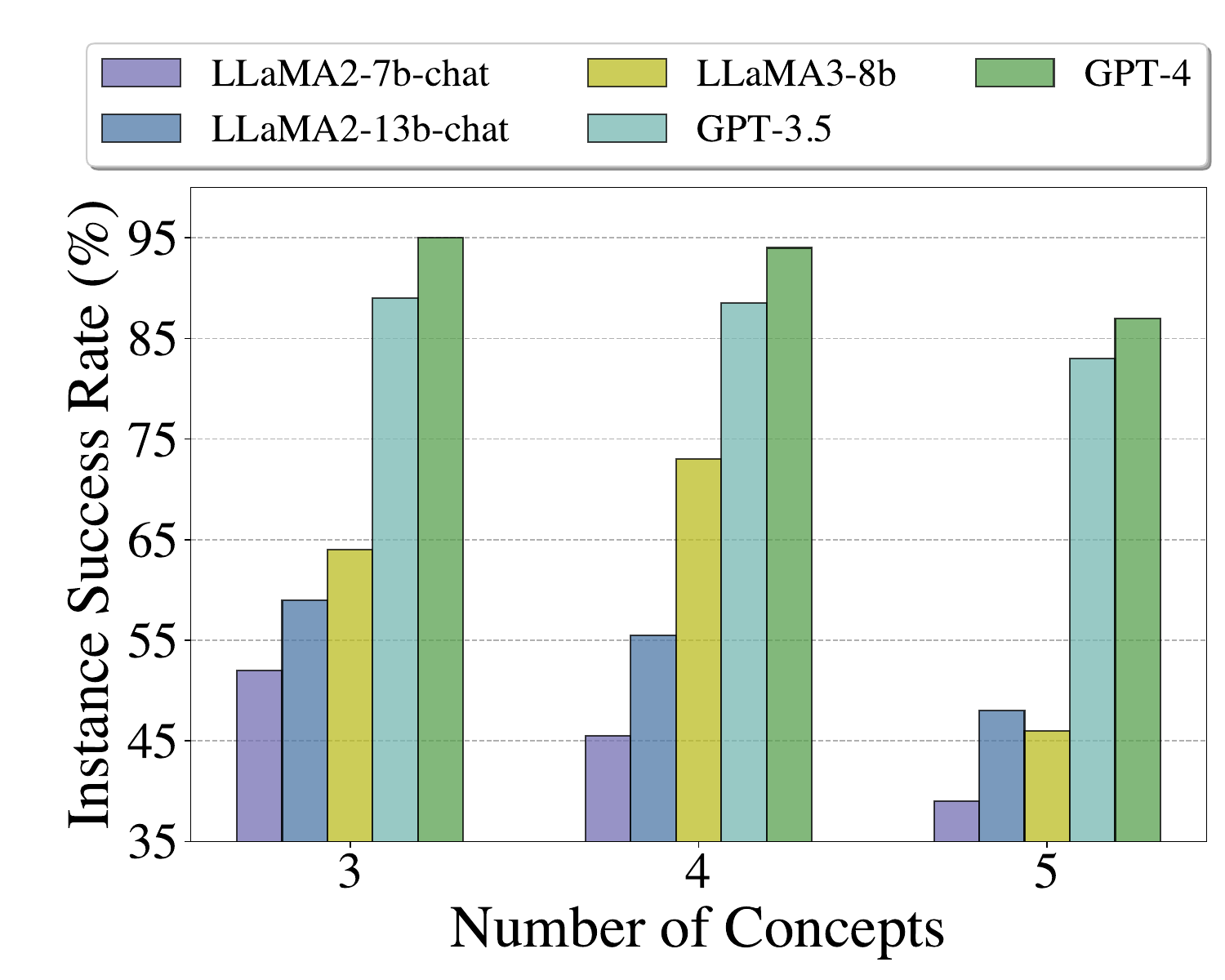}
\caption{Experimental results showing instance success rates by the number of concepts. GPT models consistently outperform LLaMA models, though performance declines across all models as the number of keywords increases.}
\label{fig:simple_constraints}
\end{figure}

We initiate our investigation with simple constraints, employing the CommonGen benchmark \cite{lin2019commongen} to assess how well LLMs generate coherent sentences from a given set of concepts.

\noindent\textbf{Experiment Setting.} CommonGen \cite{lin2019commongen}  is a constrained commonsense generation task with lexical constraints. In this experiment, we treat each concept list in CommonGen as input constraints for LLMs to generate a proper sentence.  We employ the instance success rate as the evaluation metric. 

\noindent\textbf{Evaluation Result.} Figure \ref{fig:simple_constraints} presents the results of experiments.GPT-3.5 and GPT-4 demonstrate impressive performance, achieving average instance success rates of 91\% and 95\% respectively across three distinct groups of instances. Conversely, LLaMA3-8b shows a less satisfactory average with a 63\% coverage rate, while LLaMA2-13b achieves only a 55\% rate. LLaMA2-7b records the lowest instance coverage among the evaluated models. This result suggests that the model's size significantly influences its ability to generate text that adheres to specified lexical constraints. Interestingly, LLaMA3-8b outperforms LLaMA2-13b, indicating that factors other than sheer model size may contribute to differences in model effectiveness.

\subsection{Challenging Constraints}
\label{exp: c_c}

To increase the complexity of the constraints, we expanded the number of concepts that need to be incorporated into the generated text. 

\noindent\textbf{Experiment Setting.} In this experiment, we randomly select concepts from the entire set of concepts within the CommonGen dataset to create a new, more challenging dataset. Then we repeat previous experiment setting to explore how well do LLMs adapt to increasingly complex constraints.

\begin{figure}[hb]
\centering
\includegraphics[width=\columnwidth]{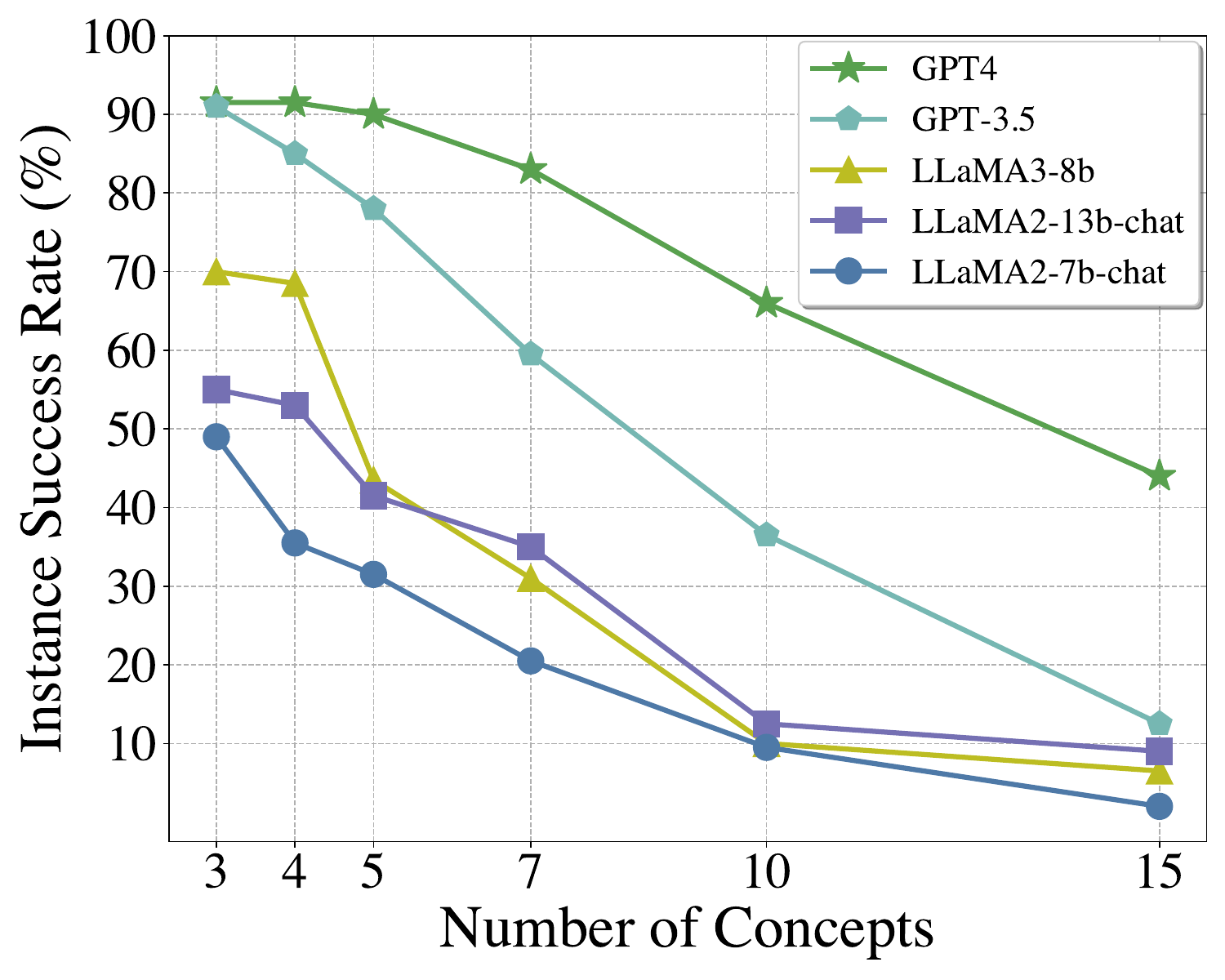}
\caption{Experiment results on instance success rate by number of keywords. As the number of keywords increases, LLMs’ performance decreases dramatically with prompt-based control.}
\label{fig:complex}
\end{figure}

\noindent\textbf{Evaluation Result.} As shown in Figure \ref{fig:complex}, there is a clear trend across all models, where the instance success rate declines as the complexity of constraints (i.e. number of concepts) increases. GPT-4 demonstrates slightly better resilience against rising complexity, maintaining a relative higher coverage rate across various groups of instances than other models. In contrast, as the number of concepts reaches 15, the performance of other models drops significantly. Notably, GPT-3.5 shows a significant decline in coverage rates; it drops from 98\% to 13\% as the number of concepts increases from 3 to 15. This sharp decrease eventually brings its performance in line with that of smaller models, such as LLaMA2-7b-chat and LLaMA2-13b-chat.

\section{Sensitive Analysis}
\label{sec:sensitive_analysis}

To better understand the factors causing LLMs to struggle with satisfying lexical constraints, we conducted a sensitivity analysis to investigate from various perspectives.

\subsection{Position Bias}

\begin{figure}[!tb]
    \centering
    \includegraphics[width=1\linewidth]{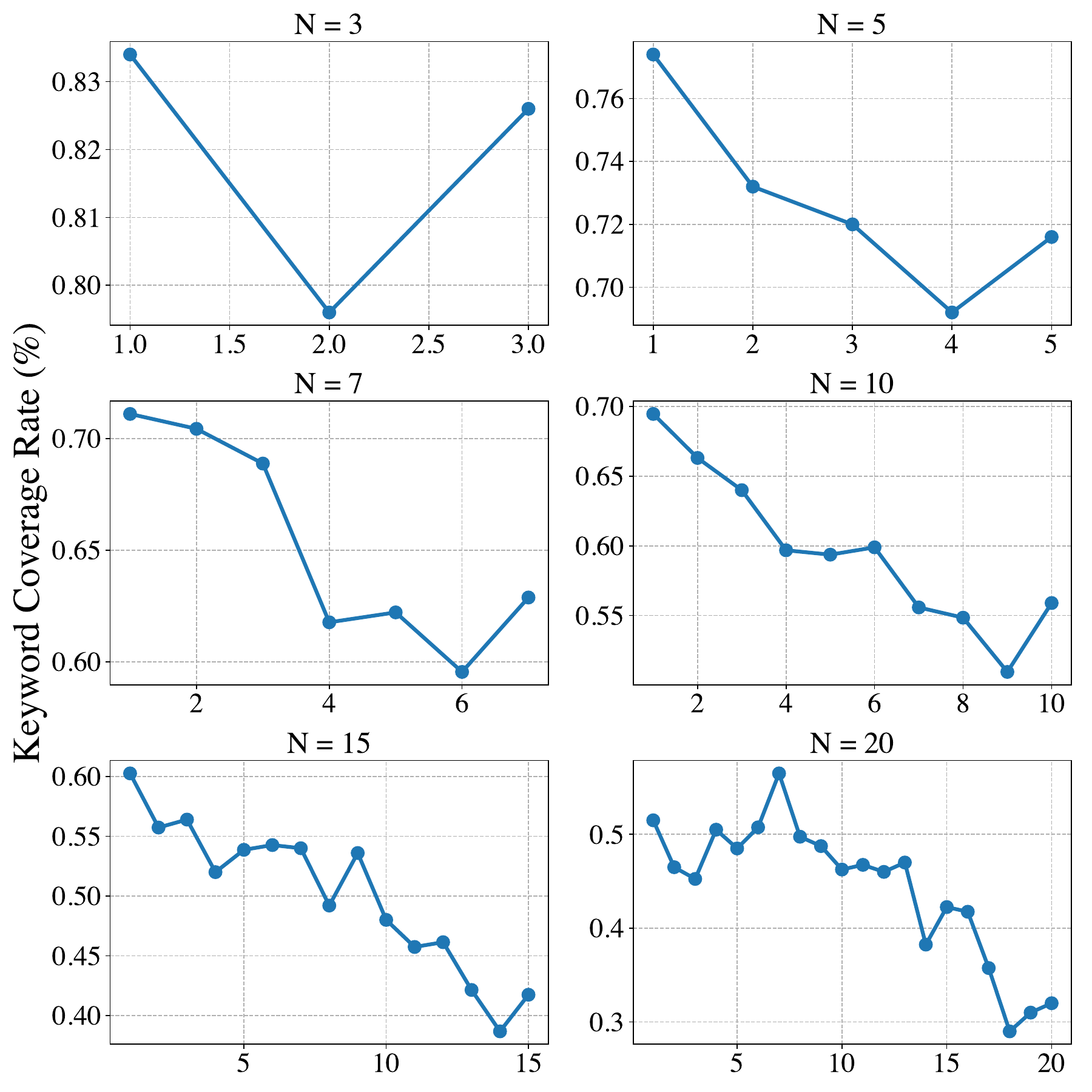}
    \caption{Experimental results on the position sensitivity of LLaMA3-8b, presented as keyword coverage rate (y-axis) for constraints placed at different positions (x-axis). Across varying numbers of keywords, the performance of LLMs exhibit similar trends.}
    \label{fig:position_bias}
    \vspace{-0.1in}
\end{figure}

The constraints are placed at varying positions within the prompt. For example, consider the prompt:
\begin{quote}
\textit{Generate a sentence with the following keywords: mountain, cat, play, jump.}
\end{quote}
Here, \textit{mountain, cat, play, jump} serve as constraints. The word "mountain" is positioned earliest in the sequence, while the word "jump" appears at the end. Previous work finds \cite{wang2023primacy} in natural language understanding tasks, wherein it tends to select labels placed at earlier positions as the answer. We aim to investigate the position bias of LLMs in LCG task.

\paragraph{Experiments Setting} We conduct experiments for 6 setting (number of keywords = [3,5,7,10,15,20]). For each setting with different specified number of keywords, we randomly select 100 sets of keywords, shuffle their positions, and conduct the experiment 20 times to ensure robustness. We evaluate the average keyword coverage rate for constraint in each position.

\paragraph{Experiment Result}
Our findings confirm that all LLMs exhibit a position bias, where keywords placed at different positions in the sequence lead to varying coverage rates. This bias is primarily attributed to either the primacy or recency effect, depending on the model. Some models, such as GPT-3.5, GPT-4, and LLaMA2-13b, are more influenced by the primacy effect, where keywords in earlier positions are more likely to be covered. Conversely, models like LLaMA2-7b and LLaMA3-8b demonstrate a stronger recency effect, prioritizing the most recently presented items. For instance, as illustrated in Figure \ref{fig:position_bias}, the keyword coverage rate decreases as the position increases from the first to the last. Keywords placed earlier in the input sequence (i.e., the prompt) are more likely to be covered than those in later positions.

This result highlights \textbf{the position of each constraint within the prompt can substantially influence the model's output}. There's the need for careful consideration of keyword placement when designing prompt for LLMs. For example, placing critical constraints in positions that are more likely to be covered can significantly enhance the effectiveness of the model in downstream tasks.

\begin{figure*}[!ht]
    \centering 
    \begin{minipage}{0.32\textwidth}
        \includegraphics[width=\linewidth]{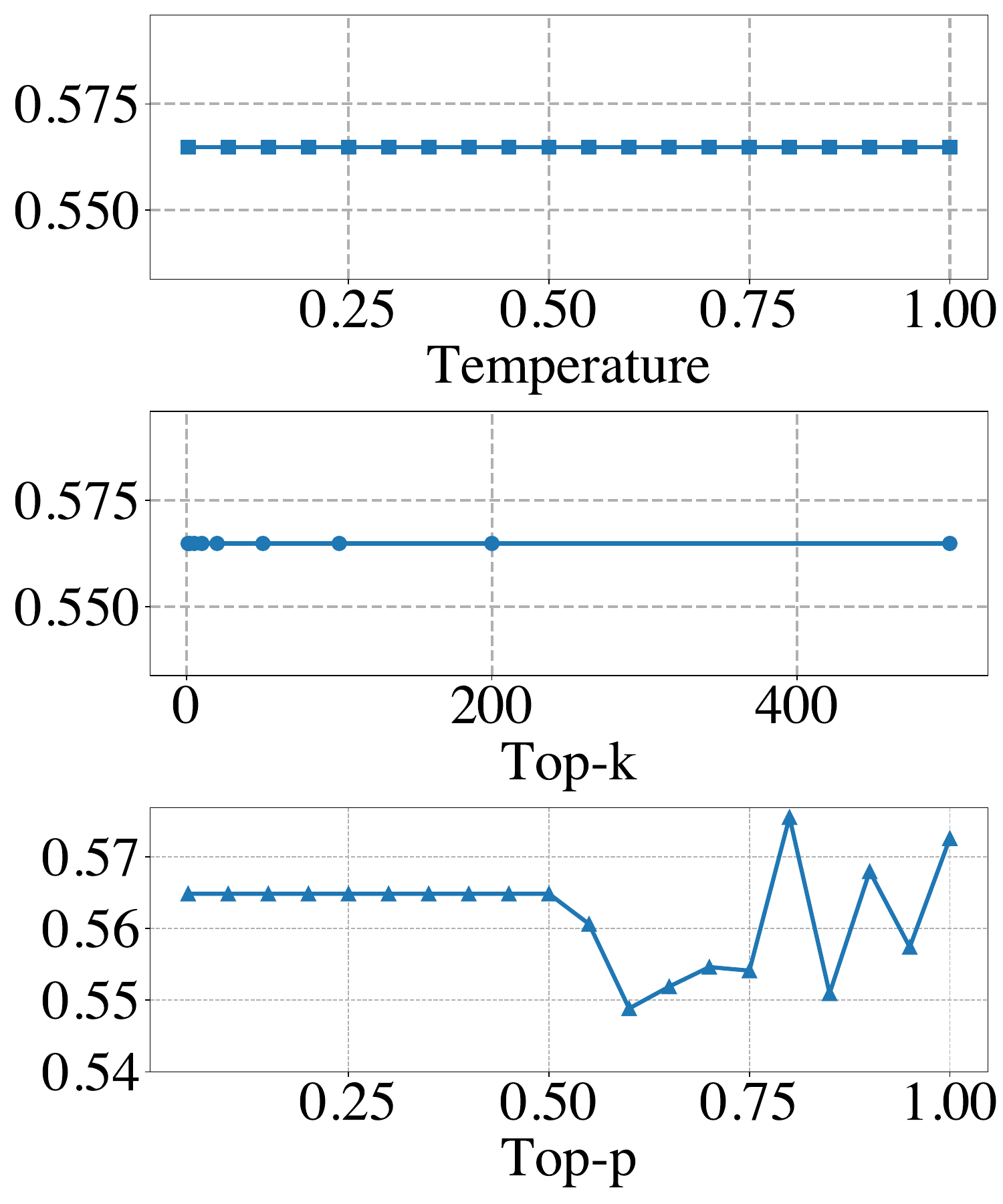}
        \caption*{(a) LLaMA2-7b-chat} 
    \end{minipage}\hfill 
    \begin{minipage}{0.32\textwidth}
        \includegraphics[width=\linewidth]{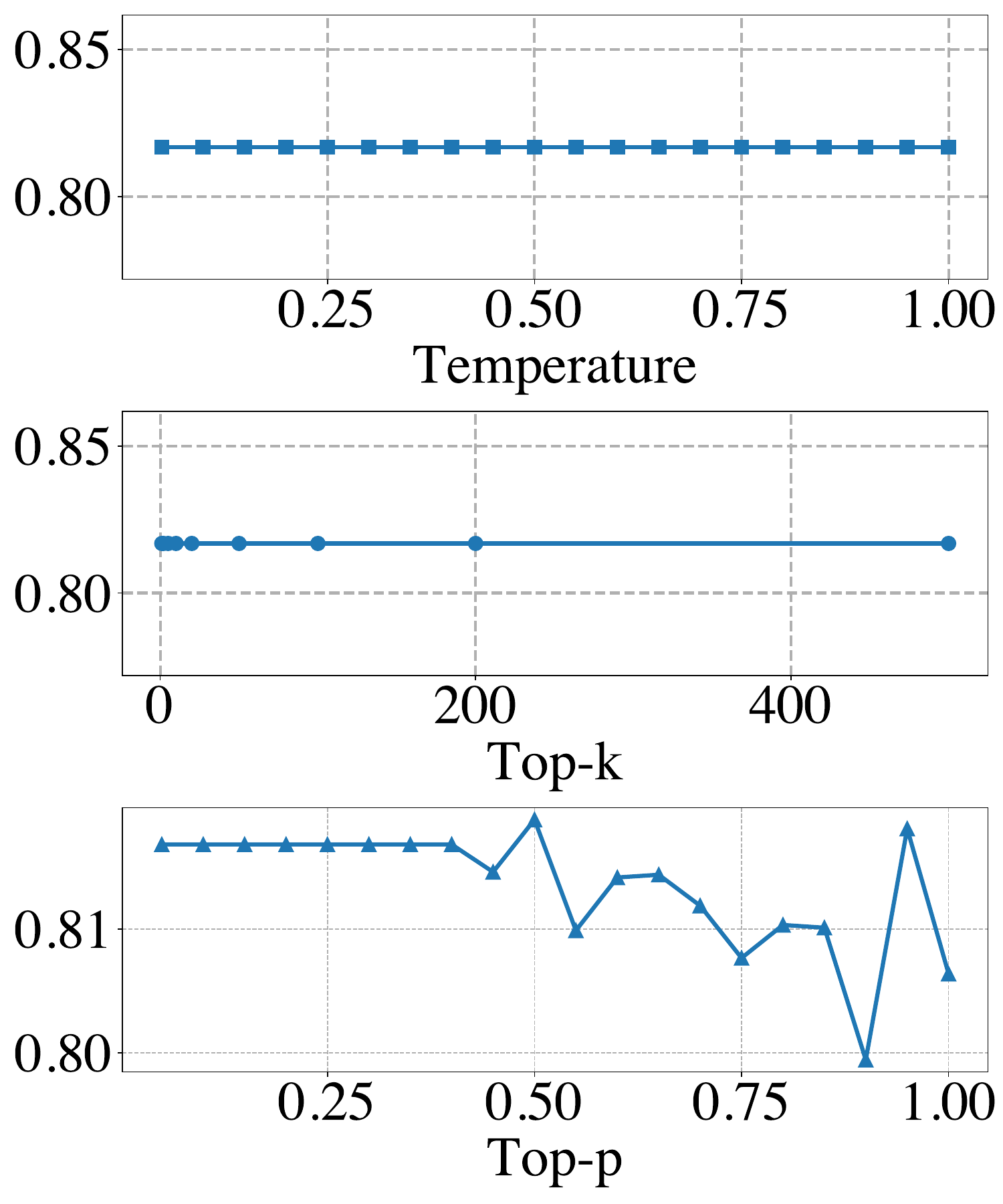}
        \caption*{(b) LLaMA2-13b-chat}
    \end{minipage}\hfill 
    \begin{minipage}{0.32\textwidth}
        \includegraphics[width=\linewidth]{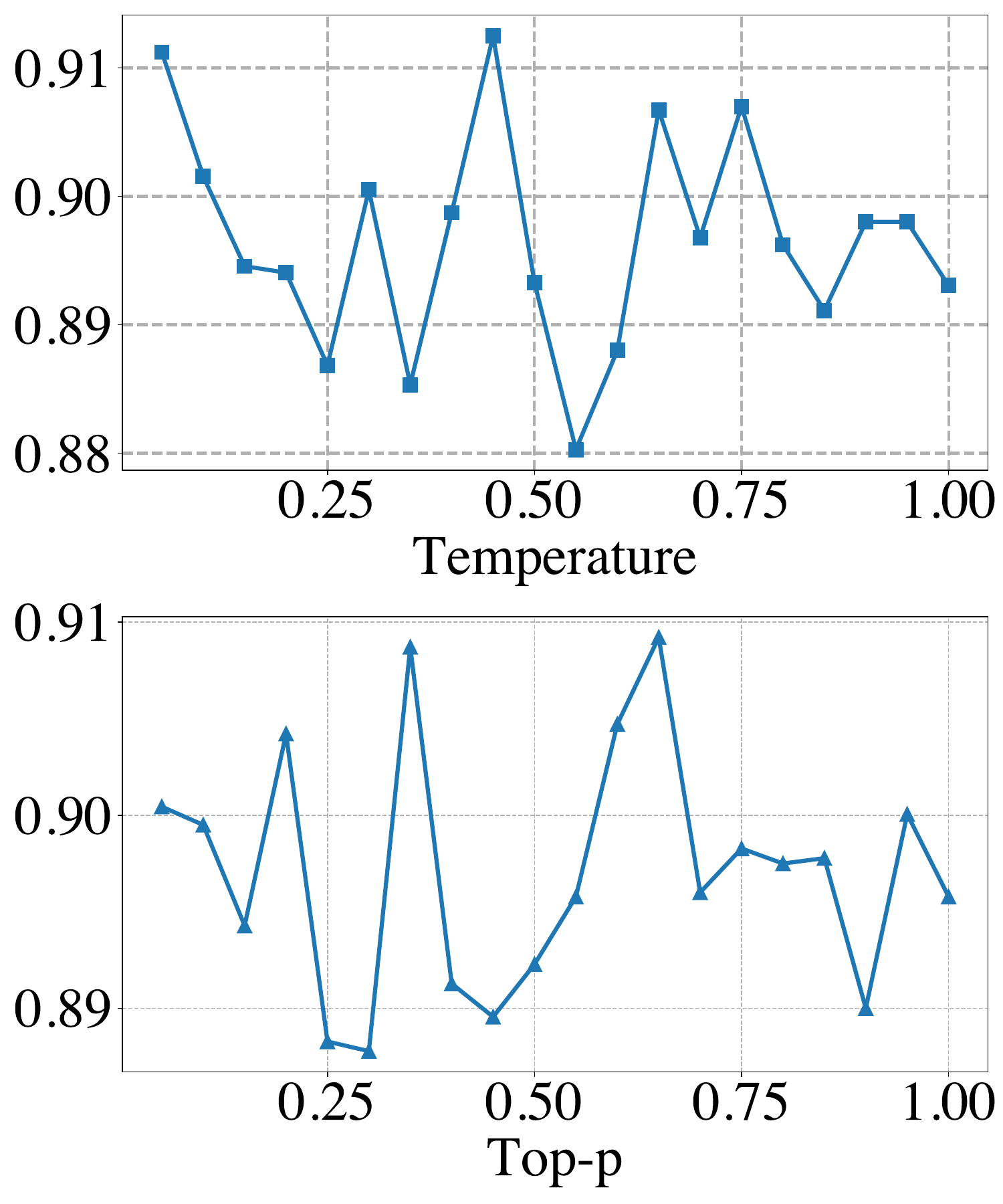}
        \caption*{(c) GPT-4}
    \end{minipage}
    \caption{Comparison of different decoding parameters across different models. The difference between the highest and the lowest scores is within 4\%, suggesting low responsiveness to control decoding parameters.} 
    \vspace{-0.1in}
    \label{fig:decoding-comparison} 
\end{figure*}

\subsection{Inherent Complexity of Compound Word}

In previous experiments on position bias, we randomly shuffled keywords to mitigate the impact of specific words on final performance. In this experiment, we isolate the position bias and investigate the effect of different keywords on the final performance.

\paragraph{Experiments Setting} From our observations in previous experiments, compound words often pose challenges in lexical processing. A compound word is formed from two or more words that collectively function as a single entity, such as "jellyfish" (a combination of "jelly" and "fish") and "anymore" (a combination of "any" and "more"). To evaluate the inherent complexity of compound words, we mixed 200 compound words with 200 random words, and conducted 5-keywords setting (i.e. generate a sentence with given five keywords) using LLaMA-13b-chat and GPT-4.

\paragraph{Experiment Result}
Our results show that LLaMA-13b-chat incorrectly split 65\% of compound words and GPT4 split 42\%, resulting in lower keyword coverage rates for compound words—35\% for LLaMA-13b-chat and 58\% for GPT4. In contrast, coverage for non-compound words was significantly higher, at 74\% for LLaMA-13b-chat and 92\% for GPT4.We can conclude that \textbf{compound words have high inherent complexity in LCG tasks}, and it's more difficult to be covered by LLMs than non-compound words. This issue could be attributed to the subword tokenization methods used by these models, which may not effectively recognize and preserve the integrity of compound words.

The separation of compound words could not only result in unsatisfied constraints, but also lead to misinterpretations or significant alterations in the intended meaning of the output. For instance, when given the task of generating a sentence using the keywords: courthouse, build, and attract, the expected outcome is a sentence related to the criminal justice system. However, LLM split 'courthouse' into 'court' and 'house'. This leads to unintended interpretations, such as generating a sentence like, "\textit{The basketball player hosted a tournament at the court built beside his house, attracting local talent to showcase their skills}." Such a sentence completely deviates from the intended context of criminal justice. 

\begin{table*}[!tb]
\small
\centering
\renewcommand{\arraystretch}{1.5}
\begin{tabular}{cccccccccccc}
\hline
\textbf{Model} &
  \multicolumn{3}{c}{\textbf{Recipe Generation}} &
   &
  \multicolumn{3}{c}{\textbf{Table to Text}} &
   &
  \multicolumn{2}{c}{\textbf{Profile Writing}} &
   \\ \hline
\rowcolor[HTML]{EFEFEF} 
 &
  \textit{n = 5} &
  \textit{n = 10} &
  \textit{n = 15} &
  \textit{} &
  \textit{n = 5} &
  \textit{n = 10} &
  \textit{n = 15} &
  \textit{} &
  \textit{n = 5} &
  \textit{n = 10} &
   \\ \cline{2-4} \cline{6-8} \cline{10-11}
LLaMA2-7b-chat  & 90\%           & 21\% & 5\%  &  & 87\%           & 21\%       & 21\% &  & 69\%       & 28\% &  \\
\rowcolor[HTML]{EFEFEF} 
LLaMA2-13b-chat & 89\%           & 27\% & 17\% &  & 84\%           & 45\%       & 39\% &  & 73\%       & 42\% &  \\
GPT-3.5         & 90\%           & 42\% & 54\% &  & {\ul 97\%}     & 80\%       & 77\% &  & 90\%       & 72\% &  \\
\rowcolor[HTML]{EFEFEF} 
GPT-4           & \textbf{100\%} & 80\% & 45\% &  & \textbf{100\%} & {\ul 87\%} & 91\% &  & {\ul 97\%} & 96\% &  \\ \hline
LLaMA2-7b-chat (DnC-5) &
  {\ul 98\%} &
  {\ul 99\%} &
  {\ul 98\%} &
   &
  \textbf{100\%} &
  \textbf{100\%} &
  {\ul 99\%} &
   &
  \textbf{100\%} &
  {\ul 99\%} &
   \\
\rowcolor[HTML]{EFEFEF} 
LLaMA2-13b-chat (DnC-5) &
  \textbf{100\%} &
  96\% &
  94\% &
   &
  \textbf{100\%} &
  \textbf{100\%} &
  \textbf{100\%} &
   &
  \textbf{100\%} &
  97\% &
   \\
GPT-3.5 (DnC-5) &
  \textbf{100\%} &
  \textbf{100\%} &
  \textbf{100\%} &
  \textbf{} &
  \textbf{100\%} &
  \textbf{100\%} &
  \textbf{100\%} &
  \textbf{} &
  \textbf{100\%} &
  \textbf{100\%} &
   \\ \hline
\end{tabular}
\caption{Results for LLMs' performance in real-word LCG task. The best results are highlighted in \textbf{boldface}, and the second-best results are \underline{underlined}. Divide and Conquer Generation strategy significantly enhances LLM performance in LCG downstream tasks.}
\vspace{-0.1in}
\label{tbl:apps}
\end{table*}

\subsection{Decoding Parameters}

We notice that LLMs are usually evaluated for LCG tasks using only default decoding parameters\cite{zhang2023survey}, or limited fixed decoding parameters \cite{sun-etal-2023-evaluating, ashok2024controllable}. We systematically varied decoding parameters to investigate the sensitivity of decoding parameters on lexical constraint generation. We aim to determine the impact of different decoding parameter settings on the performance of LLMs in LCG.

\paragraph{Experiment Setting} Follow the prior practice \cite{huang2023catastrophic}, we experiment with the following three variants for decoding strategy:
\begin{itemize}
    \item Temperature $\tau$ controls the sharpness of the next-token distribution. We vary it from $0.05$ to $1$ with step size $0.05$.
    \item Top-$K$ sampling filters the $K$ most likely next words, and then the next predicted word will be sampled among these $K$ words only. We vary $K$ in $\{1, 2, 5, 10, 20, 50, 100, 200, 500\}$.
    \item Top-$p$ sampling ~\citep{holtzman2019curious} chooses from the smallest possible set of words whose cumulative probability exceeds the probability $p$. We vary $p$ from $0.05$ to $1$ with step size $0.05$.
\end{itemize}
We evaluated all models under different decoding parameters in 10-keywords LCG task (i.e. generate sentence with given 10 keywords). Specifically, we only vary temperature and top-p parameters for GPT-3.5 and GPT-4, as we did not have control over the top-k settings.

\paragraph{Experiment Results}
Figure \ref{fig:decoding-comparison} presents the average keyword coverage rate for 150 instances, each containing 10 keywords (see Appendix \ref{sec:heatmap} for more detail). For LLaMA2-7b-chat and LLaMA2-13b-chat, there appears to be no significant effect from variations in temperature and top-k settings, and the differences observed with various top-p settings are within 4\%, suggesting a low sensitivity to the top-p parameter. While GPT-4 demonstrates more variability under different settings, the difference between the highest and lowest scores remains confined to 4\%.

This minimal variance suggests that the \textbf{decoding parameters are not highly sensitive for LLMs in LCG task, especially for temperature and top-k.}

\section{Real-world applications}
\label{sec:applications}

We have also evaluated the performance of LLMs in real-world applications to understand their practical effectiveness. In this section, we demonstrate three use cases: Recipe generation, table-to-text, and profile writing. We use the best decoding parameter configuration (\textit{Top-p} = 0.9) identified in previous section for all following experiments. Example prompt and response for each application are attached to Appendix \ref{sec:prompts}.

\subsection{Recipe Generation} 
The task is to generate a complete recipe given ingredients. LLMs need to create a coherent and structured set of cooking instructions that makes practical and culinary sense, and cover all provided keywords.

\noindent\textbf{Experiment Setting.} We randomly selected 100 food ingredients from the USDA National Nutrient Database \cite{USDA2016} and grouped them into sets with varying numbers of ingredients (n = [5, 10,15]). Each group comprises ingredients versatile enough to be applicable to multiple recipes, guaranteeing the existence of at least one valid recipe for the given combination of ingredients. LLMs is then prompted in 3-shot fashion to generate recipe with given set of ingredients, where ingredients are keywords that are expected to be contained in the generated recipe. Each generated recipe is evaluate based on the instance success rate. 

\noindent\textbf{Evaluation Result.} Table \ref{tbl:apps} presents the results of the experiment. When tasked with recipe generation, we observed that LLMs typically outline their plan in the initial sentence, such as "\textit{Lemon Garlic Pasta is quick to prepare, making it perfect for a weeknight dinner yet elegant enough for entertaining guests.}", and "\textit{To create Chicken and Mushroom Risotto, follow these steps}". \textbf{These introductory statements act as a double-edged sword}.

On the positive side, these introductory statements establish the scope for subsequent content generation, facilitating the model's ability to incorporate relevant keywords effectively. In the 5-keyword setting, the instance success rate for the LLaMA2 models increases by approximately 30\% compared to Experiment \ref{exp: c_c}, where LLMs were tasked solely with text generation under keyword constraints.

On the negative side, these introductory statements can detract from the final generation outcome if they are not accurate. If there are a large number of keywords, LLMs tend to include only a few in the first sentence, leading to the omission of the remaining keywords. As the number of keywords increases, there is a noticeable decline in performance across all models. For example, the instance success rate for LLaMA2-13b decreases from 89\% to 17\% as the number of constraints increases from 5 to 15.

\vspace{-0.1in}
\subsection{Table to Text}

Following previous work \cite{chen2023towards}, table-to-text task takes a table as input, and formulate a table as a sequence of records. We evaluate the effectiveness of LLMs in presenting the essential information from the structured data in a narrative form.

\noindent\textbf{Experiment Setting.} WIKIBIO \cite{lebret2016neural} is a dataset contain of 728,321 tables data from English Wikipedia. We processed the WIKIBIO dataset by extracting keywords from each table's column headers as ground truth, and categorizing the tables into groups based on the number of keywords identified. For each group, 200 samples are randomly selected. Next, we construct instances from each group based on number of keywords needed. LLMs is then prompted in 3-shot fashion to summarize the content of these tables in a short paragraph, and each generated summary is evaluated based on the instance success rate.

\noindent\textbf{Evaluation Result.}  As shown in table \ref{tbl:apps},GPT-4 demonstrates the strongest performance, achieving 100\% accuracy with 5 keywords setting,  and maintaining high instance success rate with larger number of keywords (87\% for n = 10 and 91\% for n = 15). However, other models, such as LLaMA2-7b-chat and LLaMA2-13b-chat, show notable declines in accuracy as the sample size increases, with significant drops from 87\% to 21\% and from 84\% to 39\%. This result indicates that LLMs struggle in satisfying more nuanced and complex constraints.

\subsection{Profile Writing}
Profile writing provides a quick overview of the client's basic information, significantly impacting decision-making and enhancing operational effectiveness. For instance, in healthcare, profiles summarize patient histories to guide treatment plans; in finance, they help assess client risk and customize financial services;  and in the legal field, detailed client profiles are crucial for informed case management. This process can be viewed as a lexical constraint generation task, where the client's information acts as the constraint, and the resulting profile paragraph serves as the output. 

\noindent\textbf{Experiment Setting.} This task is aimed to generate a profile contain all specific features of a client. We obtained data consists of various attributes of clients to assessing risk score \footnote{We conduct experiment on an open-source dataset: \url{https://www.kaggle.com/datasets/parisrohan/credit-score-classification}}, such as age, employment details, education, housing level, etc. In our experiment, we extract individual client information from this dataset, and prompt LLMs to generate a detailed profile graph contain all information.

\noindent\textbf{Evaluation Result.} Table \ref{tbl:apps} presents the results of the experiment. Similar to previous experiments, GPT-4 demonstrates the highest consistency and robustness among the models, scoring 97\% with n = 5 and 96\% with n = 10, showing only a slight decrease in performance with an increase in number of constraints. Other models show more significant drops in performance, denoting the need of improvement strategy.

\section{Divide and Conquer Generation}

As demonstrated in previous experiments, LLMs face significant challenges in satisfying increasingly complex constraints. To address these difficulties, we propose a simple and effective strategy—Divide and Conquer Generation (DnC) —to improve LLMs' performance in Language Constraint Generation (LCG), which suitble for both white-box and black-box models.

\begin{algorithm}[!tb]
\caption{Divide and Conquer Generation}
\label{algo:dnc}
\begin{algorithmic}[1]
\State \textbf{Input:} Set of keywords $X$, maximum number of iterations $k$
\State \textbf{Output:} Output sentence $s$, initially empty.

\While{$X$ is not empty}
    \State $s' \gets \text{generate a sentence using } X$
    \State $Y \gets \text{extract words from } s'$
    \State $s \gets \text{merge } s' \text{ with } s$
    \State $X \gets X \setminus Y$ \Comment{remove included keywords}
    
    \If{$\text{number of iterations exceeds } k$}
        \State \Return $s$
    \EndIf
\EndWhile
\end{algorithmic}
\end{algorithm}

\subsection{Method}

From our observation, we found LLMs struggle with complex tasks that encompass a large amount of keywords. In contrast, they exhibit a high success rate when dealing with simpler tasks involving a smaller number of keywords, which motivate us to break down the complex task to several simple tasks in divide and conquer fashion. 

Algorithm \ref{algo:dnc} illustrates DnC strategy. Recall that the task is to generate a natural sentence containing the token sequence \( Y = [y_1, y_2, \ldots, y_N] \) using a specified set of \( N \) keywords \( X = [x_1, x_2, \ldots, x_N] \), such that \( X \subseteq Y \). Our strategy iteratively generates sentences while addressing the missing keywords \( X \setminus Y = \{ x \in X \mid x \notin Y \} \) from each generation iteration, then merge these sentences into a cohesive final output. Figure \ref{fig:teaser} contains detailed example of the process of our strategy. We repeat this process until all constraints are satisfied, or exceed the max allowed number of iteration $K$.

\subsection{Performance Evaluation}

Rejection Sampling (RJ) is a Monte Carlo algorithm to sample data from a sophisticated  distribution with the help of a proxy distribution \cite{DBLP:books/sp/RobertC04}. This method can assist with black-box models, where texts that do not meet certain criteria are discarded, and the sampling process is iteratively repeated. We choose rejection sampling as the baseline method, and evaluate the DnC strategy.

\begin{figure}[!tb]
    \centering
    \includegraphics[width=1\linewidth]{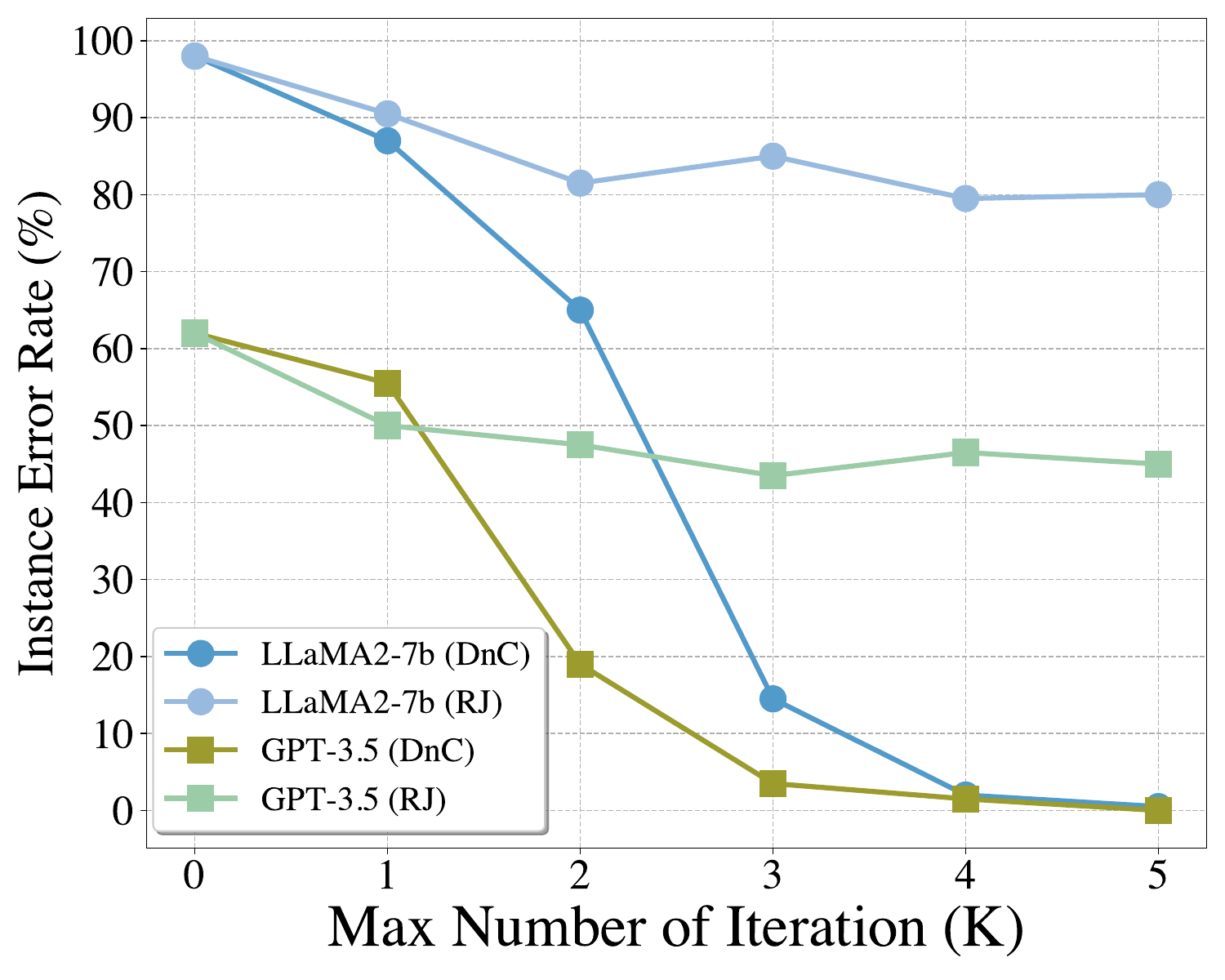}
    \caption{Comparison experiment between Rejection Sampling (RJ) and Divide-and-Conquer Generation (DnC). The x-axis represents the maximum number of allowed iterations, while the y-axis shows the error rate for each approach in satisfying all lexical constraints. The Divide-and-Conquer Generation strategy improves LLM performance by over 40\% compared to the baseline.}
    \label{fig:cost}
    \vspace{-0.1in}
\end{figure}

We repeat the 15-keyword generation experiment with LLaMA2-7b-chat and GPT-3.5, using both RJ and  DnC strategy under varying maximum number of iterations $K$ allowed. Figure \ref{fig:cost} demonstrate the result, where y-axis is the error rate in satisfying all lexical constraints (i.e. 1 minus the instance success rate). At $K = 0$, the models generate in a vanilla setting, without employing any specific strategies. From the result, we can observe that  while the RJ strategy manages to reduce the error rate, it does not lead to significant improvements. In the contrast,  DnC help both model achieve a near-perfect performance (error rate close to 0\%) with $ K = 4$. With the help of DnC, LLaMA2-7b-chat model decrease error rate from approximately 96\% to 3\%, demonstrating the effectiveness of the DnC approach.

Furthermore, we revisited application tasks introduced in \ref{sec:applications}.
Table \ref{tbl:apps} compares the instance success rates for each approach. From the result, with the implementation of the DnC strategy, all models achieve near-perfect performance (instance coverage rates approaching 100\%). Specifically, the LLaMA2-7b-chat model records an average improvement of 61\% across all tasks with the help of DnC strategy. Notably, GPT-3.5 (DNC-5) achieves a 100\% instance success rate for all tasks.

\subsection{Quality Evaluation}
The DnC strategy assumes that LLMs can generate fluent and high-quality text. To evaluate this, we conducted additional experiments to assess the overall quality of the text produced using the DnC strategy. We randomly selected 20 samples from both the vanilla and Divide-and-Conquer generation strategy (DnC-5) for the GPT-4o models, and performed both human and automatic evaluations as outlined below.

First, we recruited 5 volunteers to manually evaluate the readability of the generated text. Each volunteer rated the text on a scale from 1 to 5, considering factors of coherence, fluency, and readability. The inter-annotator agreement was approximately 0.8 based on the Pearson correlation metric. The result demonstrates that the DnC-5 strategy delivers comparable performance to the vanilla models. The readability and fluency of the generated text, both with and without Divide and Conquer, receive scores of 5.0, indicating high readability. The average coherence score of the generated text of DnC-5 strategy is 4.88, only slightly lower than generated text without DnC-5 strategy (4.94), demonstrating the overall high quality generated text.

Second, we utilized GPT-4-turbo-0409 to automatically evaluate the text quality. We prompted GPT-4-turbo-0409 to evaluate the generated text quality in a one-shot fashion. Specifically, we provided one example from the human evaluation and then asked GPT-4-turbo-0409 to rate the input text. The prompt used here is provided in Appendix \ref{app:prompt_eval}. The results indicate that the DnC-5 strategy performs comparably to the vanilla generation strategy, with only minimal differences. Specifically, GPT-4o with DnC-5 achieves nearly identical average scores as GPT-4o vanilla in coherence (4.9 vs. 5.0) and fluency (4.9 vs. 5.0), and the same perfect readability score (5.0).

These evaluations ensure that our model not only satisfies lexical constraints but also produces high-quality and readable text.

\section{Related Work}

\paragraph{LLMs Evaluation on Controllable Generation}
With recent advancements in Large Language Models (LLMs), there is increasing interest in evaluating controllable text generation tasks \citep{liu-etal-2024-benchmarking, sun-etal-2023-evaluating,  zhang2023survey, ashok2024controllable}. For example, \newcite{sun-etal-2023-evaluating} conducted evaluations of these tasks and discovered that LLMs often struggle to meet fine-grained constraints. However, their analysis of lexical constraint generation was limited to relatively simple constraints in a narrow context. Our work expands on this by conducting a more comprehensive and in-depth analysis of lexical constraint generation, providing deeper insights into the capabilities and limitations of LLMs in this area.

\paragraph{Lexically Constrained Generation}
There are many works trying to improve lexically constrained generation. We roughly categorize these studies: (1) Fine tuning:  In auto-regressive models, controlling over the generated contents can be naturally achieved by fine-tuning or retraining the models on examples with specific attributes  ~\citep{peng2018towards,keskar2019ctrl}. However, these methods are considered expensive in inference and low quality in generated texts. (2) Post processing: Injecting constraints into the decoding algorithm is one approach to addressing LCG tasks. Methods like constrained beam search \citep{anderson2017guided, post2018fast, hokamp-liu-2017-lexically} and NeuroLogic decoding \citep{lu2021neurologic, lu2022neurologic} follow this strategy. Another approach involves using an auxiliary discriminator to guide the model based on the desired attribute, as proposed by PPLM \citep{dathathri2020pplm}. Methods such as GeDi \citep{krause2021gedi} and DEXPERTS \citep{liu2021dexpert} employ contrastive learning by training an auxiliary language model to adjust the token distribution at each step. Though shown effectiveness in lexically constrained generation, these algorithmic methods are not suitable for recent pre-trained LLMs due to the black-box nature.  (2) Specialized model structure: InsNET is an expressive insertion-based text generator with efficient training and flexible decoding~\cite{lu2022insnet}. (3) Prompt-based control methods: Another line of research focuses on improving lexically constrained generation through prompt-based control mechanisms \citep{sheng2020towards,shin2020autoprompt,letster2021power,li2021prefix,iso2022autotemplate}. However, these methods often exhibit weaker controllability and have struggled to demonstrate effectiveness in real-world applications involving modern LLMs. 

\section {Conclusion}

We systematically conduct in-depth analysis on LLMs in satisfying lexical constraints, and identify the current challenges faced by LLMs in satisfying lexical constraints, including position bias, low responsiveness to control decoding parameters, and struggle with handling the inherent complexity of certain constraints (e.g. compound word). To tackle these challenge, we provide a effective novel solution, Divide and Conquer Generation strategy, paving the way for more sophisticated downstream applications.

\section*{Limitation}

Our work is not without limitations. First, our experiments are prompt-based, requiring extensive prompt engineering effort. While we selected the best-performing prompts available, there remains the possibility that more effective prompts could further enhance the reported result. Second, automatic evaluations have inherent imperfections. Third, the proposed Divide and Conquer (DnC) strategy increases the number of API calls, resulting in higher costs. Additionally, due to the nature of the DnC approach, the coherence of the merged results may be compromised. We leave the evaluation and further refinement of this to future work.

\section*{Acknowledgement}
The work is partially support by a DARPA ANSR program FA8750-23-2-0004 and a National Science Foundation CAREER award \#2339766, and a research award NSF \#2331966. 
The views and conclusions are those of the authors and should not reflect the official policy or position of DARPA or the U.S. Government.

\newpage
\bibliography{custom}

\begin{thebibliography}{46}
\providecommand{\natexlab}[1]{#1}

\bibitem[{Achiam et~al.(2023)Achiam, Adler, Agarwal, Ahmad, Akkaya, Aleman, Almeida, Altenschmidt, Altman, Anadkat et~al.}]{achiam2023gpt}
Josh Achiam, Steven Adler, Sandhini Agarwal, Lama Ahmad, Ilge Akkaya, Florencia~Leoni Aleman, Diogo Almeida, Janko Altenschmidt, Sam Altman, Shyamal Anadkat, et~al. 2023.
\newblock Gpt-4 technical report.
\newblock \emph{arXiv preprint arXiv:2303.08774}.

\bibitem[{Anderson et~al.(2017)Anderson, Fernando, Johnson, and Gould}]{anderson2017guided}
Peter Anderson, Basura Fernando, Mark Johnson, and Stephen Gould. 2017.
\newblock Guided open vocabulary image captioning with constrained beam search.
\newblock In \emph{{EMNLP}}.

\bibitem[{Ashok and Poczos(2024)}]{ashok2024controllable}
Dhananjay Ashok and Barnabas Poczos. 2024.
\newblock Controllable text generation in the instruction-tuning era.
\newblock \emph{arXiv preprint arXiv:2405.01490}.

\bibitem[{Brown et~al.(2020)Brown, Mann, Ryder, Subbiah, Kaplan, Dhariwal, Neelakantan, Shyam, Sastry, Askell et~al.}]{brown2020language}
Tom Brown, Benjamin Mann, Nick Ryder, Melanie Subbiah, Jared~D Kaplan, Prafulla Dhariwal, Arvind Neelakantan, Pranav Shyam, Girish Sastry, Amanda Askell, et~al. 2020.
\newblock Language models are few-shot learners.
\newblock \emph{Advances in neural information processing systems}, 33:1877--1901.

\bibitem[{Chen et~al.(2022)Chen, Lu, Xu, Li, Jingbo, Dou, and Xiong}]{chen2023towards}
Miao Chen, Xinjiang Lu, Tong Xu, Yanyan Li, Zhou Jingbo, Dejing Dou, and Hui Xiong. 2022.
\newblock \href {https://doi.org/10.18653/v1/2022.emnlp-main.562} {Towards table-to-text generation with pretrained language model: A table structure understanding and text deliberating approach}.
\newblock In \emph{Proceedings of the 2022 Conference on Empirical Methods in Natural Language Processing}, pages 8199--8210, Abu Dhabi, United Arab Emirates. Association for Computational Linguistics.

\bibitem[{Dathathri et~al.(2020)Dathathri, Madotto, Lan, Hung, Frank, Molino, Yosinski, and Liu}]{dathathri2020pplm}
Sumanth Dathathri, Andrea Madotto, Janice Lan, Jane Hung, Eric Frank, Piero Molino, Jason Yosinski, and Rosanne Liu. 2020.
\newblock Plug and play language models: {A} simple approach to controlled text generation.
\newblock In \emph{{ICLR}}. OpenReview.net.

\bibitem[{Goldfarb-Tarrant et~al.(2020)Goldfarb-Tarrant, Chakrabarty, Weischedel, and Peng}]{goldfarb2020content}
Seraphina Goldfarb-Tarrant, Tuhin Chakrabarty, Ralph Weischedel, and Nanyun Peng. 2020.
\newblock Content planning for neural story generation with aristotelian rescoring.
\newblock In \emph{the 2020 Conference on Empirical Methods in Natural Language Processing (EMNLP)}, pages 4319--4338.

\bibitem[{H.~Lee et~al.(2020)H.~Lee, Shu, Achananuparp, Prasetyo, Liu, Lim, and Varshney}]{h2020recipegpt}
Helena H.~Lee, Ke~Shu, Palakorn Achananuparp, Philips~Kokoh Prasetyo, Yue Liu, Ee-Peng Lim, and Lav~R Varshney. 2020.
\newblock Recipegpt: Generative pre-training based cooking recipe generation and evaluation system.
\newblock In \emph{Companion Proceedings of the Web Conference 2020}, pages 181--184.

\bibitem[{Hokamp and Liu(2017)}]{hokamp-liu-2017-lexically}
Chris Hokamp and Qun Liu. 2017.
\newblock \href {https://doi.org/10.18653/v1/P17-1141} {Lexically constrained decoding for sequence generation using grid beam search}.
\newblock In \emph{Proceedings of the 55th Annual Meeting of the Association for Computational Linguistics (Volume 1: Long Papers)}, pages 1535--1546, Vancouver, Canada. Association for Computational Linguistics.

\bibitem[{Holtzman et~al.(2019)Holtzman, Buys, Du, Forbes, and Choi}]{holtzman2019curious}
Ari Holtzman, Jan Buys, Li~Du, Maxwell Forbes, and Yejin Choi. 2019.
\newblock The curious case of neural text degeneration.
\newblock In \emph{International Conference on Learning Representations}.

\bibitem[{Huang et~al.(2024)Huang, Gupta, Xia, Li, and Chen}]{huang2023catastrophic}
Yangsibo Huang, Samyak Gupta, Mengzhou Xia, Kai Li, and Danqi Chen. 2024.
\newblock Catastrophic jailbreak of open-source llms via exploiting generation.
\newblock In \emph{The Twelfth International Conference on Learning Representations}.

\bibitem[{Iso(2024)}]{iso2022autotemplate}
Hayate Iso. 2024.
\newblock \href {https://aclanthology.org/2024.inlg-main.1} {{A}uto{T}emplate: A simple recipe for lexically constrained text generation}.
\newblock In \emph{Proceedings of the 17th International Natural Language Generation Conference}, pages 1--12, Tokyo, Japan. Association for Computational Linguistics.

\bibitem[{Keskar et~al.(2019)Keskar, McCann, Varshney, Xiong, and Socher}]{keskar2019ctrl}
Nitish~Shirish Keskar, Bryan McCann, Lav~R. Varshney, Caiming Xiong, and Richard Socher. 2019.
\newblock {CTRL:} {A} conditional transformer language model for controllable generation.
\newblock \emph{CoRR}, abs/1909.05858.

\bibitem[{Krause et~al.(2021)Krause, Gotmare, McCann, Keskar, Joty, Socher, and Rajani}]{krause2021gedi}
Ben Krause, Akhilesh~Deepak Gotmare, Bryan McCann, Nitish~Shirish Keskar, Shafiq~R. Joty, Richard Socher, and Nazneen~Fatema Rajani. 2021.
\newblock Gedi: Generative discriminator guided sequence generation.
\newblock In \emph{{EMNLP} (Findings)}, pages 4929--4952. Association for Computational Linguistics.

\bibitem[{Lebret et~al.(2016)Lebret, Grangier, and Auli}]{lebret2016neural}
R{\'e}mi Lebret, David Grangier, and Michael Auli. 2016.
\newblock \href {https://doi.org/10.18653/v1/D16-1128} {Neural text generation from structured data with application to the biography domain}.
\newblock In \emph{Proceedings of the 2016 Conference on Empirical Methods in Natural Language Processing}, pages 1203--1213, Austin, Texas. Association for Computational Linguistics.

\bibitem[{Lester et~al.(2021)Lester, Al{-}Rfou, and Constant}]{letster2021power}
Brian Lester, Rami Al{-}Rfou, and Noah Constant. 2021.
\newblock The power of scale for parameter-efficient prompt tuning.
\newblock In \emph{{EMNLP} {(1)}}.

\bibitem[{Li and Liang(2021)}]{li2021prefix}
Xiang~Lisa Li and Percy Liang. 2021.
\newblock Prefix-tuning: Optimizing continuous prompts for generation.
\newblock In \emph{{ACL/IJCNLP} {(1)}}.

\bibitem[{Lin et~al.(2020)Lin, Zhou, Shen, Zhou, Bhagavatula, Choi, and Ren}]{lin2019commongen}
Bill~Yuchen Lin, Wangchunshu Zhou, Ming Shen, Pei Zhou, Chandra Bhagavatula, Yejin Choi, and Xiang Ren. 2020.
\newblock \href {https://doi.org/10.18653/v1/2020.findings-emnlp.165} {{C}ommon{G}en: A constrained text generation challenge for generative commonsense reasoning}.
\newblock In \emph{Findings of the Association for Computational Linguistics: EMNLP 2020}, pages 1823--1840, Online. Association for Computational Linguistics.

\bibitem[{Liu et~al.(2021)Liu, Sap, Lu, Swayamdipta, Bhagavatula, Smith, and Choi}]{liu2021dexpert}
Alisa Liu, Maarten Sap, Ximing Lu, Swabha Swayamdipta, Chandra Bhagavatula, Noah~A. Smith, and Yejin Choi. 2021.
\newblock Dexperts: Decoding-time controlled text generation with experts and anti-experts.
\newblock In \emph{{ACL/IJCNLP} {(1)}}.

\bibitem[{Liu et~al.(2024)Liu, Fabbri, Chen, Zhao, Han, Joty, Liu, Radev, Wu, and Cohan}]{liu-etal-2024-benchmarking}
Yixin Liu, Alexander Fabbri, Jiawen Chen, Yilun Zhao, Simeng Han, Shafiq Joty, Pengfei Liu, Dragomir Radev, Chien-Sheng Wu, and Arman Cohan. 2024.
\newblock \href {https://doi.org/10.18653/v1/2024.findings-naacl.280} {Benchmarking generation and evaluation capabilities of large language models for instruction controllable summarization}.
\newblock In \emph{Findings of the Association for Computational Linguistics: NAACL 2024}, pages 4481--4501, Mexico City, Mexico. Association for Computational Linguistics.

\bibitem[{Lu et~al.(2022{\natexlab{a}})Lu, Meng, and Peng}]{lu2022insnet}
Sidi Lu, Tao Meng, and Nanyun Peng. 2022{\natexlab{a}}.
\newblock Insnet: An efficient, flexible, and performant insertion-based text generation model.
\newblock \emph{Advances in Neural Information Processing Systems}, 35:7011--7023.

\bibitem[{Lu et~al.(2024)Lu, Zhao, Tao, Gupta, Wu, Chung, and Peng}]{lu2024nado2}
Sidi Lu, Wenbo Zhao, Chenyang Tao, Arpit Gupta, Shanchan Wu, Tagyoung Chung, and Nanyun Peng. 2024.
\newblock Dinado: Norm-disentangled neurally-decomposed oracles for controlling language models.
\newblock In \emph{Proceedings of the Fortieth International Conference on Machine Learning (ICML)}.

\bibitem[{Lu et~al.(2022{\natexlab{b}})Lu, Welleck, West, Jiang, Kasai, Khashabi, Bras, Qin, Yu, Zellers, Smith, and Choi}]{lu2022neurologic}
Ximing Lu, Sean Welleck, Peter West, Liwei Jiang, Jungo Kasai, Daniel Khashabi, Ronan~Le Bras, Lianhui Qin, Youngjae Yu, Rowan Zellers, Noah~A. Smith, and Yejin Choi. 2022{\natexlab{b}}.
\newblock Neurologic a*esque decoding: Constrained text generation with lookahead heuristics.
\newblock In \emph{{NAACL-HLT}}.

\bibitem[{Lu et~al.(2022{\natexlab{c}})Lu, Welleck, West, Jiang, Kasai, Khashabi, Le~Bras, Qin, Yu, Zellers, Smith, and Choi}]{lu2021neurologic}
Ximing Lu, Sean Welleck, Peter West, Liwei Jiang, Jungo Kasai, Daniel Khashabi, Ronan Le~Bras, Lianhui Qin, Youngjae Yu, Rowan Zellers, Noah~A. Smith, and Yejin Choi. 2022{\natexlab{c}}.
\newblock \href {https://doi.org/10.18653/v1/2022.naacl-main.57} {{N}euro{L}ogic a*esque decoding: Constrained text generation with lookahead heuristics}.
\newblock In \emph{Proceedings of the 2022 Conference of the North American Chapter of the Association for Computational Linguistics: Human Language Technologies}, pages 780--799, Seattle, United States. Association for Computational Linguistics.

\bibitem[{Meng et~al.(2022)Meng, Lu, Peng, and Chang}]{meng2022controllable}
Tao Meng, Sidi Lu, Nanyun Peng, and Kai-Wei Chang. 2022.
\newblock Controllable text generation with neurally-decomposed oracle.
\newblock \emph{Advances in Neural Information Processing Systems}, 35:28125--28139.

\bibitem[{Meta(2024)}]{metaMetaLlama}
Meta. 2024.
\newblock {M}eta {L}lama 3 --- llama.meta.com.
\newblock \url{https://llama.meta.com/llama3/}.
\newblock [Accessed 16-06-2024].

\bibitem[{Ouyang et~al.(2022)Ouyang, Wu, Jiang, Almeida, Wainwright, Mishkin, Zhang, Agarwal, Slama, Ray et~al.}]{ouyang2022training}
Long Ouyang, Jeffrey Wu, Xu~Jiang, Diogo Almeida, Carroll Wainwright, Pamela Mishkin, Chong Zhang, Sandhini Agarwal, Katarina Slama, Alex Ray, et~al. 2022.
\newblock Training language models to follow instructions with human feedback.
\newblock \emph{Advances in neural information processing systems}, 35:27730--27744.

\bibitem[{Peng et~al.(2018)Peng, Ghazvininejad, May, and Knight}]{peng2018towards}
Nanyun Peng, Marjan Ghazvininejad, Jonathan May, and Kevin Knight. 2018.
\newblock Towards controllable story generation.
\newblock In \emph{NAACL Workshop}.

\bibitem[{Post and Vilar(2018)}]{post2018fast}
Matt Post and David Vilar. 2018.
\newblock Fast lexically constrained decoding with dynamic beam allocation for neural machine translation.
\newblock In \emph{{NAACL-HLT}}.

\bibitem[{Qian et~al.(2022)Qian, Dong, Shen, Wei, and Chen}]{qian2022controllable}
Jing Qian, Li~Dong, Yelong Shen, Furu Wei, and Weizhu Chen. 2022.
\newblock \href {https://doi.org/10.18653/v1/2022.findings-acl.229} {Controllable natural language generation with contrastive prefixes}.
\newblock In \emph{Findings of the Association for Computational Linguistics: ACL 2022}, pages 2912--2924, Dublin, Ireland. Association for Computational Linguistics.

\bibitem[{Robert and Casella(2004)}]{DBLP:books/sp/RobertC04}
Christian~P. Robert and George Casella. 2004.
\newblock \href {https://doi.org/10.1007/978-1-4757-4145-2} {\emph{Monte Carlo Statistical Methods}}.
\newblock Springer Texts in Statistics. Springer.

\bibitem[{Sha(2020)}]{sha-2020-gradient}
Lei Sha. 2020.
\newblock \href {https://doi.org/10.18653/v1/2020.emnlp-main.701} {Gradient-guided unsupervised lexically constrained text generation}.
\newblock In \emph{Proceedings of the 2020 Conference on Empirical Methods in Natural Language Processing (EMNLP)}, pages 8692--8703, Online. Association for Computational Linguistics.

\bibitem[{Sheng et~al.(2020)Sheng, Chang, Natarajan, and Peng}]{sheng2020towards}
Emily Sheng, Kai-Wei Chang, Premkumar Natarajan, and Nanyun Peng. 2020.
\newblock Towards controllable biases in language generation.
\newblock In \emph{{EMNLP}(Findings)}.

\bibitem[{Shin et~al.(2020)Shin, Razeghi, IV, Wallace, and Singh}]{shin2020autoprompt}
Taylor Shin, Yasaman Razeghi, Robert L.~Logan IV, Eric Wallace, and Sameer Singh. 2020.
\newblock Autoprompt: Eliciting knowledge from language models with automatically generated prompts.
\newblock In \emph{{EMNLP} {(1)}}.

\bibitem[{Sun et~al.(2023)Sun, Tian, Zhou, Xu, Hu, Gupta, Wieting, Peng, and Ma}]{sun-etal-2023-evaluating}
Jiao Sun, Yufei Tian, Wangchunshu Zhou, Nan Xu, Qian Hu, Rahul Gupta, John Wieting, Nanyun Peng, and Xuezhe Ma. 2023.
\newblock \href {https://doi.org/10.18653/v1/2023.emnlp-main.190} {Evaluating large language models on controlled generation tasks}.
\newblock In \emph{Proceedings of the 2023 Conference on Empirical Methods in Natural Language Processing}, pages 3155--3168, Singapore. Association for Computational Linguistics.

\bibitem[{Tian and Peng(2022)}]{tian2022sonnet}
Yufei Tian and Nanyun Peng. 2022.
\newblock Zero-shot sonnet generation with discourse-level planning and aesthetics features.
\newblock In \emph{2022 Annual Conference of the North American Chapter of the Association for Computational Linguistics (NAACL)}.

\bibitem[{Touvron et~al.(2023)Touvron, Martin, Stone, Albert, Almahairi, Babaei, Bashlykov, Batra, Bhargava, Bhosale et~al.}]{touvron2023llama}
Hugo Touvron, Louis Martin, Kevin Stone, Peter Albert, Amjad Almahairi, Yasmine Babaei, Nikolay Bashlykov, Soumya Batra, Prajjwal Bhargava, Shruti Bhosale, et~al. 2023.
\newblock Llama 2: Open foundation and fine-tuned chat models.
\newblock \emph{arXiv preprint arXiv:2307.09288}.

\bibitem[{{US Department of Agriculture, Agricultural Research Service}(2016)}]{USDA2016}
{US Department of Agriculture, Agricultural Research Service}. 2016.
\newblock Usda national nutrient database for standard reference, release 28 (slightly revised).
\newblock \url{http://www.ars.usda.gov/nea/bhnrc/mafcl}.
\newblock Version Current: May 2016.

\bibitem[{Wang et~al.(2023)Wang, Cai, Chen, Liang, and Hooi}]{wang2023primacy}
Yiwei Wang, Yujun Cai, Muhao Chen, Yuxuan Liang, and Bryan Hooi. 2023.
\newblock \href {https://doi.org/10.18653/v1/2023.emnlp-main.8} {Primacy effect of {C}hat{GPT}}.
\newblock In \emph{Proceedings of the 2023 Conference on Empirical Methods in Natural Language Processing}, pages 108--115, Singapore. Association for Computational Linguistics.

\bibitem[{Yang et~al.(2023)Yang, Liu, Lei, Yang, Xue, Chen, and Xie}]{yang2022tailor}
Kexin Yang, Dayiheng Liu, Wenqiang Lei, Baosong Yang, Mingfeng Xue, Boxing Chen, and Jun Xie. 2023.
\newblock \href {https://doi.org/10.18653/v1/2023.acl-long.25} {Tailor: A soft-prompt-based approach to attribute-based controlled text generation}.
\newblock In \emph{Proceedings of the 61st Annual Meeting of the Association for Computational Linguistics (Volume 1: Long Papers)}, pages 410--427, Toronto, Canada. Association for Computational Linguistics.

\bibitem[{Yao et~al.(2019)Yao, Peng, Ralph, Knight, Zhao, and Yan}]{yao2019plan}
Lili Yao, Nanyun Peng, Weischedel Ralph, Kevin Knight, Dongyan Zhao, and Rui Yan. 2019.
\newblock Plan-and-write: Towards better automatic storytelling.
\newblock In \emph{The Thirty-Third AAAI Conference on Artificial Intelligence (AAAI-19)}.

\bibitem[{Zhang et~al.(2023{\natexlab{a}})Zhang, Song, Li, Zhou, and Song}]{zhang2023survey}
Hanqing Zhang, Haolin Song, Shaoyu Li, Ming Zhou, and Dawei Song. 2023{\natexlab{a}}.
\newblock A survey of controllable text generation using transformer-based pre-trained language models.
\newblock \emph{ACM Computing Surveys}, 56(3):1--37.

\bibitem[{Zhang et~al.(2023{\natexlab{b}})Zhang, Dang, Peng, and Broeck}]{zhang2023gelato}
Honghua Zhang, Meihua Dang, Nanyun Peng, and Guy Van~den Broeck. 2023{\natexlab{b}}.
\newblock Tractable control for autoregressive language generation.
\newblock In \emph{Proceedings of the Fortieth International Conference on Machine Learning (ICML)}.

\bibitem[{Zhang et~al.(2024)Zhang, Kung, Yoshida, den Broeck, and Peng}]{zhang2024adaptable}
Honghua Zhang, Po-Nien Kung, Masahiro Yoshida, Guy~Van den Broeck, and Nanyun Peng. 2024.
\newblock Adaptable logical control for large language models.
\newblock In \emph{Proceedings of The Thirty-eighth Annual Conference on Neural Information Processing Systems (NeurIPS)}.

\bibitem[{Zhang et~al.(2023{\natexlab{c}})Zhang, Dong, Li, Zhang, Sun, Wang, Li, Hu, Zhang, Wu et~al.}]{zhang2023instruction}
Shengyu Zhang, Linfeng Dong, Xiaoya Li, Sen Zhang, Xiaofei Sun, Shuhe Wang, Jiwei Li, Runyi Hu, Tianwei Zhang, Fei Wu, et~al. 2023{\natexlab{c}}.
\newblock Instruction tuning for large language models: A survey.
\newblock \emph{arXiv preprint arXiv:2308.10792}.

\bibitem[{Zhou et~al.(2023)Zhou, Jiang, Wilcox, Cotterell, and Sachan}]{zhou2023controlled}
Wangchunshu Zhou, Yuchen~Eleanor Jiang, Ethan Wilcox, Ryan Cotterell, and Mrinmaya Sachan. 2023.
\newblock Controlled text generation with natural language instructions.
\newblock In \emph{International Conference on Machine Learning}, pages 42602--42613. PMLR.

\end{thebibliography}

\clearpage
\appendix

\section{Prompts and Example Response}
\label{sec:prompts}

Prompt and example response for lexical constrained generation, and downstream applications.

\subsection{Example of Lexical Constrained generation}
\paragraph{Prompt} Generate a sentence using the following keywords: cat, leaves, sunny, leaps, energy

\paragraph{Response} Generate a sentence using the following keywords: cat, leaves, sunny, leaps, energy

\subsection{Example of Recipe Generation}
\paragraph{Prompt} Generate a recipe using the following ingredients: milk, beef,
rice.
\paragraph{Response} The recipe for Creamy Beef and Rice Casserole: Begin by cooking 1 cup of rice according to package instructions and set aside to cool. Add
1 pound of ground beef to the skillet, breaking it apart with
a spatula. Cook until the beef is thoroughly browned and no
pink remains. Drain any excess fat and set aside.In a large
casserole dish, combine the cooked rice, browned beef, and 2
cups of milk, mixing well. Stir in 1 cup of shredded cheddar
cheese, season with salt and freshly ground black pepper to taste,
and add a dash of paprika for a smoky flavor, if desired. Dot
the top of the mixture with small pieces of butter to enrich the
flavor.Preheat your oven to 350°F (175°C). Cover the casserole
dish with aluminum foil and bake for 25 minutes. Remove the
foil and continue baking for an additional 5-10 minutes, or until
the top is bubbly and golden brown. Remove from the oven and
let it sit for a few minutes to settle

\subsection{Example of Table-to-text}
\paragraph{Prompt} Generate a short paragraph to summarize the table
content: \\

\begin{tabular}{|l|l|}
\hline
\textbf{Nationality} & German       \\ \hline
\textbf{Birth Date}  & 1954         \\ \hline
\textbf{Name}        & Walter Extra \\ \hline
\textbf{\begin{tabular}[c]{@{}l@{}}Article\\ Title\end{tabular}} & \begin{tabular}[c]{@{}l@{}}Aviation Legend:\\ Walter Extra\end{tabular}       \\ \hline
\textbf{Occupation}                                              & \begin{tabular}[c]{@{}l@{}}Aircraft Designer \\ and Manufacturer\end{tabular} \\ \hline
\end{tabular}
\\

\paragraph{Response} Walter extra is a german award-winning aerobatic
pilot, chief aircraft designer and founder of extra
flugzeugbau (extra aircraft construction )), a manufacturer of aerobatic aircraft . extra was trained as a mechanical engineer. Walter extra has designed a series of performance aircraft which include unlimited aerobatic aircraft and turboprop transports.

\subsection{Example of Profile Writing}
\paragraph{Prompt} Write a profiling paragraph contains all features of the
client: Name: Ben Smith; Age: 29; Gender: Male; Race: White; Marriage: Single; FICO:740; Education: Bachelor; Occupation: Engineer; Location: Chicago; Housing: Rent.
\paragraph{Response} Ben Smith is a 29-year-old white male, currently single, residing in Chicago. He holds a Bachelor's degree and works as an engineer, a profession that demands high technical proficiency and strong problem-solving skills. Ben rents his housing, which could suggest flexibility in his living arrangements or a preference for less permanent commitments in terms of residency. With a FICO score of 740, Ben demonstrates strong financial responsibility, indicating good management of credit and other financial obligations. This profile suggests that Ben is both stable in his career and adaptable in his personal life.

\subsection{Prompt used for automatic evaluation}
\label{app:prompt_eval}
\begin{Verbatim}[frame=single]
Please evaluate the following text 
based on three criteria: coherence, 
fluency, and overall readability. 
Provide a score from 1 to 5 for 
each criterion, 
along with a brief explanation 
for your rating. 
Use the following definitions 
for the criteria:

- Coherence: How logically consistent 
  and well-organized the text is.
- Fluency: How smoothly and naturally 
  the text reads.
- Readability: How easy the text is 
  to read and understand.

Example Evaluation: [examples here]

Text to Evaluate:
"[Text to be evaluate here]"

Please provide detailed feedback 
on each criterion to help improve 
the quality of the text generation.
\end{Verbatim}

\newpage
\begin{minipage}{\textwidth}
\section{Decoding Hyperparameters}
\label{sec:heatmap}
We visualize the instance success rate for each instance under different decoding hyperparameters. Heat-maps are attached as following.

\begin{figure}[H]
    \centering
    \includegraphics[width=\linewidth]{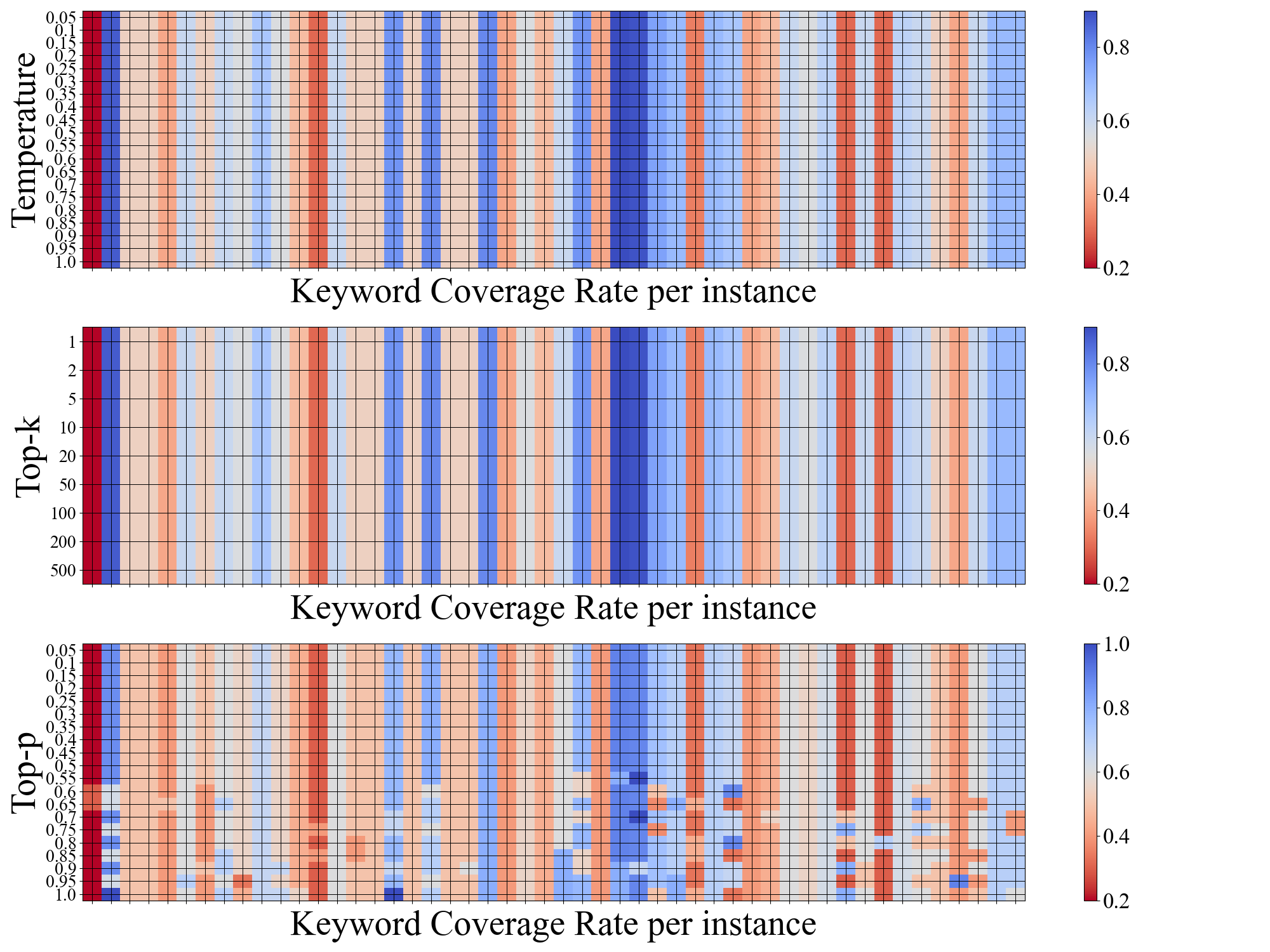}
    \caption{Heatmap of keyword coverage rate per instance for LLaMA2-7b model.}
\end{figure}
\end{minipage}



\begin{figure*}
    \centering
    \includegraphics[width=1.05\linewidth]{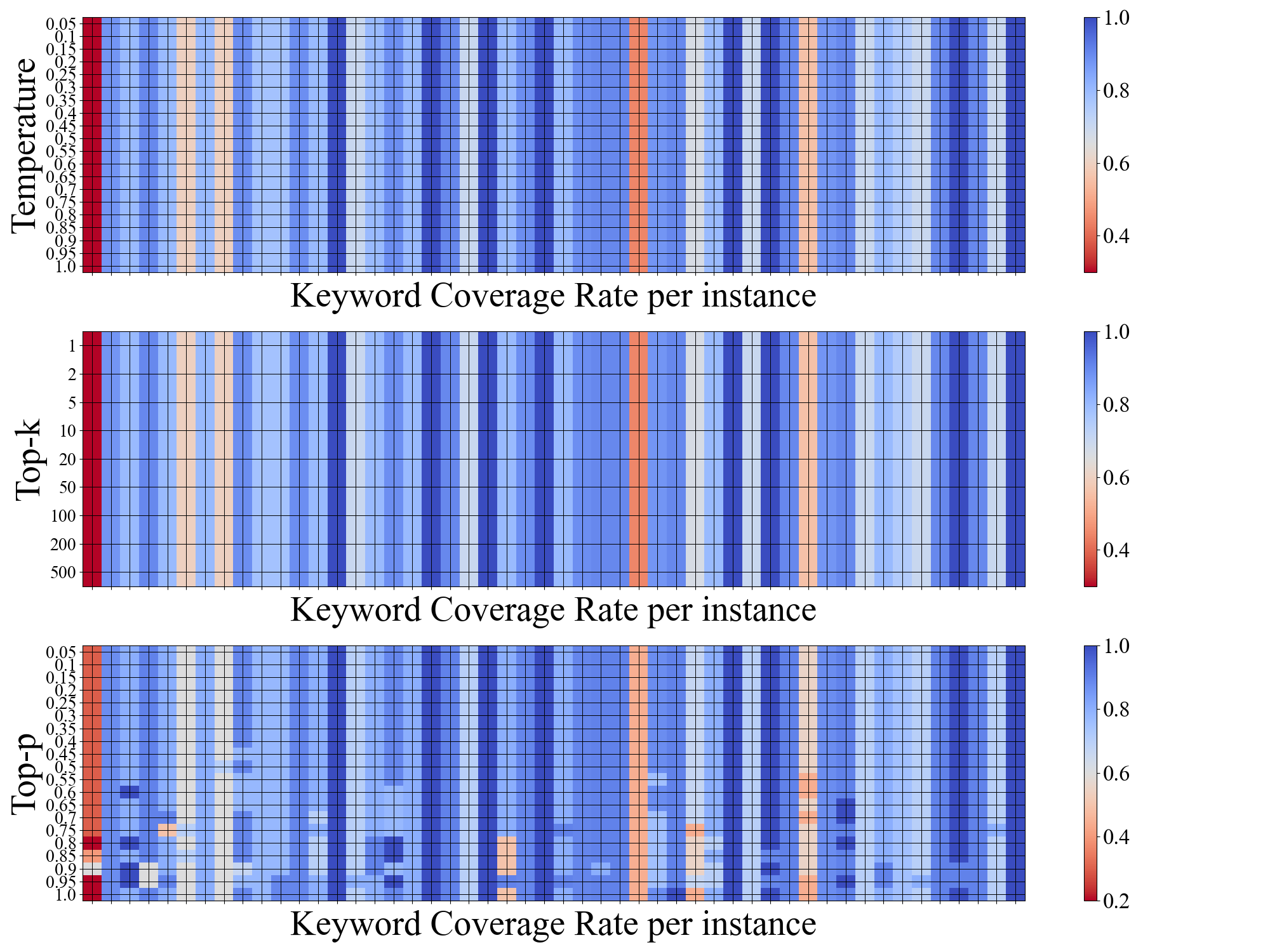}
    \caption{Heatmap of keyword coverage rate per instance for LLaMA2-13b model. }
\end{figure*}

\begin{figure*}
    \centering
    \includegraphics[width=1.05\linewidth]{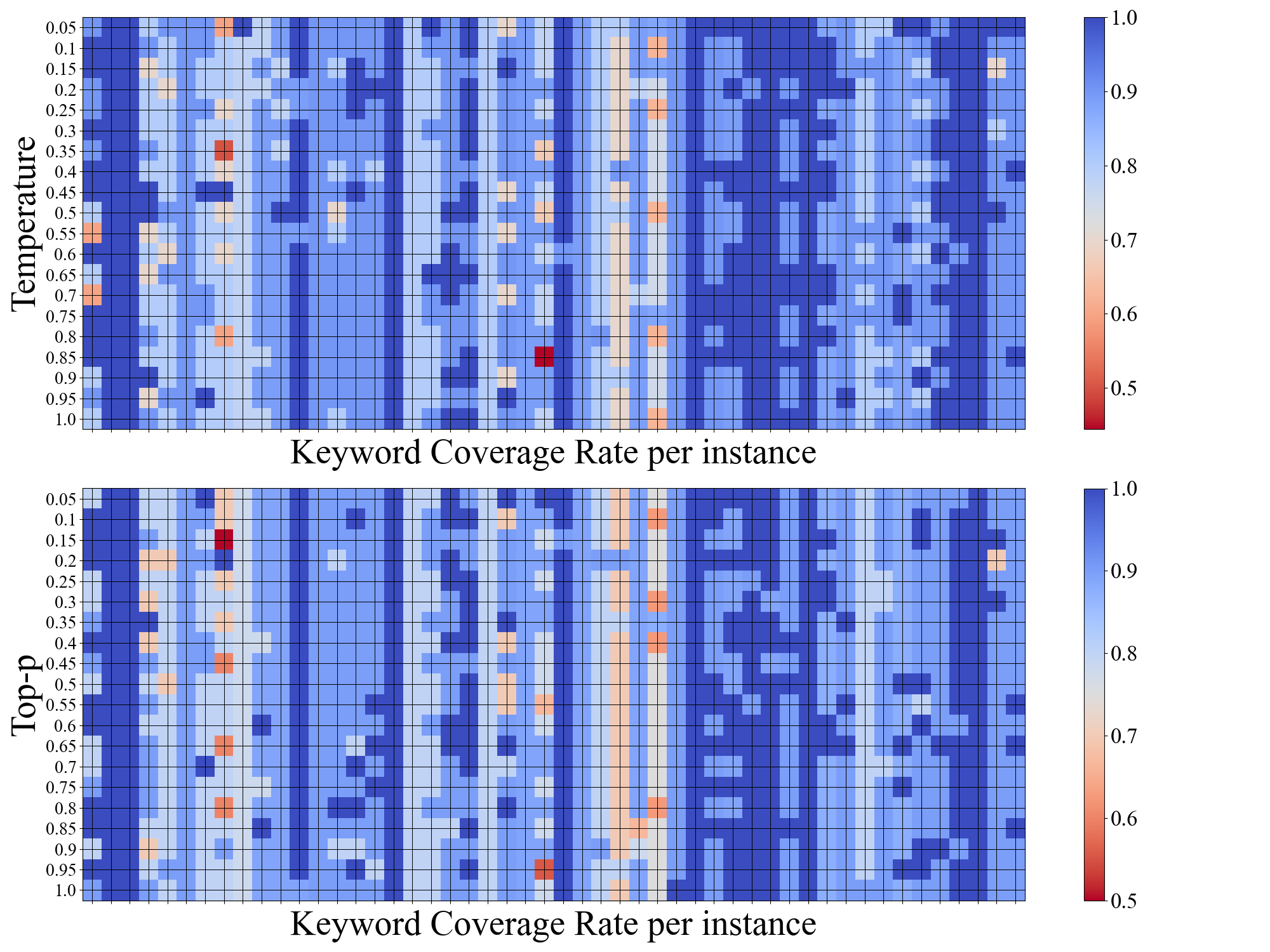}
    \caption{Heatmap of keyword coverage rate per instance for GPT model. }
\end{figure*}

\begin{figure*}
    \centering
    \includegraphics[width=1.05\linewidth]{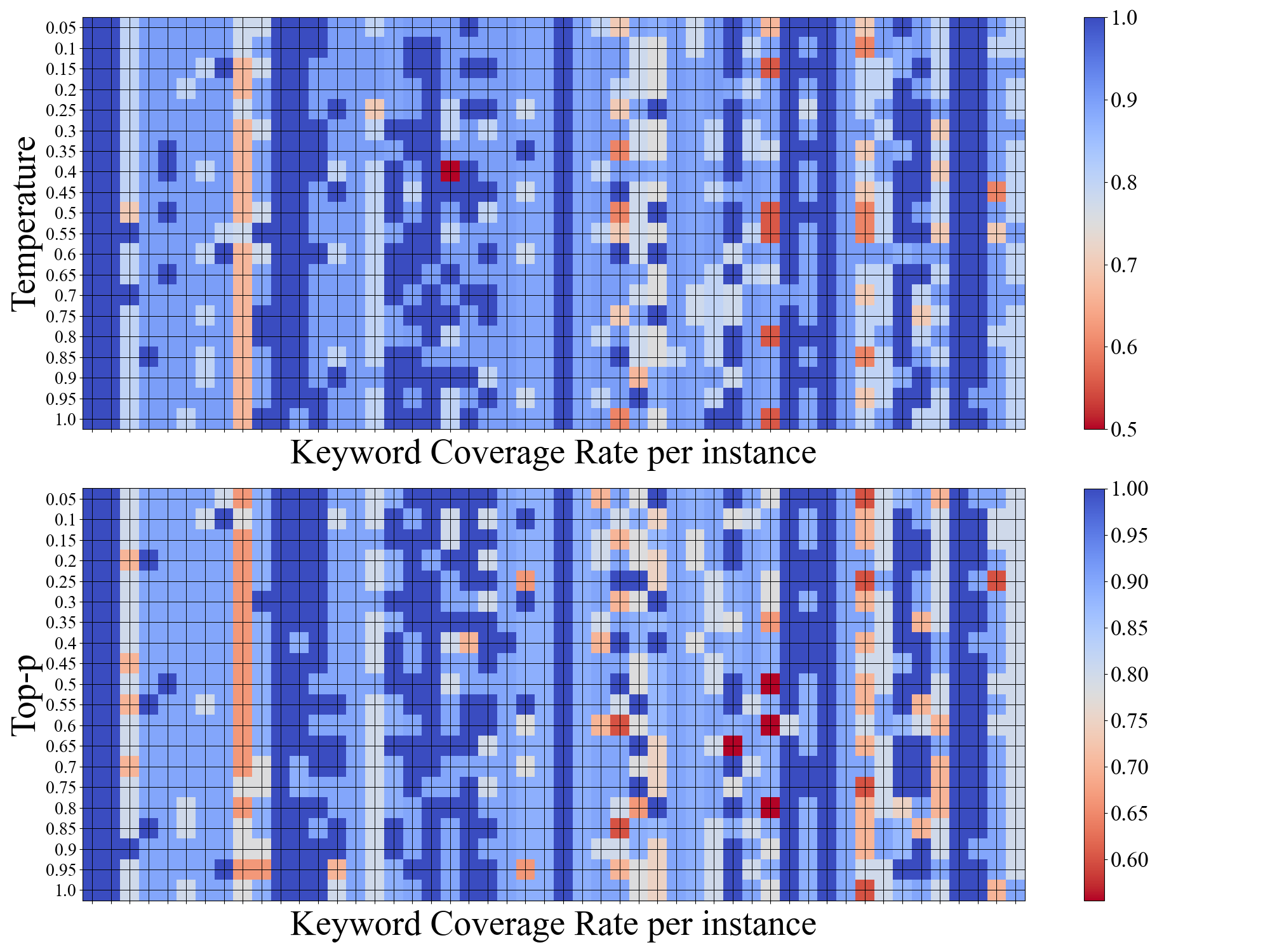}
    \caption{Heatmap of keyword coverage rate per instance for GPT-4 model. }
\end{figure*}

\end{document}